%% file: TensorRingCompletion.tex
\documentclass[journal]{IEEEtran}

\usepackage{times}
\usepackage{epsfig}
\usepackage{graphicx}
\usepackage{amsmath}
\usepackage{amssymb}
\input{wwq_package}

\usepackage{hyperref}

\usepackage{cite}

\begin{document}

\title{Efficient Low Rank Tensor Ring Completion}

\author{Wenqi Wang, Vaneet Aggarwal, and Shuchin Aeron \thanks{W. Wang and V. Aggarwal are with Purdue University, West Lafayette IN 47907, email: \{wang2041, vaneet\}@purdue.edu. S. Aeron is with Tufts University, Medford, MA 02155‎, email:  shuchin@ece.tufts.edu. }}

\maketitle

\input{abstractintro}

\input{NotationPreliminaries}
\input{Algorithm}

\input{Complexity}
\input{Experiment}

\input{conclusion}

\input{Apdix}

\bibliographystyle{IEEEtran}
\bibliography{Ref}

\end{document}

%% file: wwq_package.tex
\usepackage{color}
\usepackage[mathscr]{eucal}
\usepackage{amsbsy}
\usepackage{bm}
\usepackage{fixltx2e}
\MakeRobust{\overrightarrow}
\usepackage{booktabs}
\usepackage{mathtools}

\usepackage{amssymb}
\usepackage{amsmath,amsthm}
\usepackage{graphicx}
\usepackage{graphics}
\usepackage{color}
\usepackage{xspace}
\usepackage{bbm}
\usepackage{psfrag}
\usepackage{algorithmicx}
\usepackage{algorithm}
\usepackage{algpseudocode}
\usepackage{multirow}
\usepackage{array}
\usepackage{url}
\usepackage[normalem]{ulem}
\usepackage{floatflt,setspace}
\usepackage{algcompatible}
\newtheorem{definition}{Definition}
\newtheorem{lemma}{Lemma}
\newtheorem{remark}{Remark}

\usepackage{stackengine}

\algnewcommand\INPUT{\item[\textbf{Input:}]}%
\algnewcommand\OUTPUT{\item[\textbf{Output:}]}%

\usepackage{graphicx}
\usepackage{caption}
\usepackage{subcaption}
\usepackage{amsmath}
\usepackage{amssymb}
\usepackage{array}
\usepackage{booktabs}

\newcommand{\argmin}{{\text{argmin}}}

%% file: abstractintro.tex
\begin{abstract}
Using the matrix product state (MPS) representation
of the recently proposed tensor ring decompositions, in this paper we propose a tensor
completion algorithm, which is an alternating minimization algorithm
 that alternates
over the factors
in the MPS representation. This development is motivated in part
by the success of matrix completion algorithms that alternate
over the (low-rank) factors. In this paper, we propose a spectral initialization for the tensor ring completion algorithm and 
 analyze the computational complexity of the proposed algorithm. 
We numerically compare it
with existing methods that employ a low rank tensor train approximation
for data completion and show that our method outperforms
the existing ones for a variety of real computer vision settings,
 and thus demonstrate the improved expressive power of tensor ring as compared to tensor train.
\end{abstract}

\section{Introduction}
Tensor decompositions for representing and storing data have recently 
 attracted considerable attention
due to their effectiveness in compressing data for statistical signal processing   \cite{SIREV,CichockiMPCZZL14,tensorface1,novikov2015tensorizing,ashraphijuo2016deterministic}. 
In this paper we focus on Tensor Ring (TR) decomposition \cite{Zhao2016} and in particular its relation to Matrix Product States (MPS) \cite{orus2014practical} representation for tensor representation and use it for completing data from missing entries. In this context our algorithm is motivated by recent work in matrix completion where under a suitable initialization an alternating minimization algorithm \cite{jain2013low, Hardt13a} over the low rank factors is able to accurately predict the missing data. 

Recently, tensor networks, considered as the generalization
of tensor decompositions, have emerged as the potentially
powerful tools for analysis of large-scale tensor
data \cite{orus2014practical}. 
The most popular tensor
network is the Tensor Train (TT) representation, which for a order-$d$ tensor with each dimension of size $n$ requires $O(dnr^2)$ parameters, where $r$ is the rank of each of the factors, and thus
allows for the efficient data representation \cite{oseledets2011tensor}.  
Tensor completion based on tensor train decompositions have been recently considered in \cite{grasedyck2015variants,phien2016efficient}. The authors of \cite{grasedyck2015variants} considered the completion of data based on the alternating least square method.

Although the TT format has been widely applied in numerical analysis, its applications to image classification and completion are rather limited \cite{novikov2015tensorizing,grasedyck2015variants,phien2016efficient}. 
As outlined in \cite{Zhao2016}, TT decomposition suffers from the following limitations. Namely, (i) TT model requires rank-1 constraints on the border factors, (ii) TT ranks are typically small for near-border factors and large for the middle factors, and (iii) the multiplications of the TT factors are not permutation invariant.
In order to alleviate those drawbacks, a tensor ring (TR) decomposition  has been proposed in \cite{Zhao2016}. TR decomposition removes the unit rank constraints for the boundary tensor factors and utilizes a trace operation in the decomposition. 
The multilinear products between cores also have no strict ordering and the cores can be circularly shifted  due to the properties of the trace operation. This  paper provides novel algorithms for data completion when the data is modeled as a TR decomposition.

For data completion using tensor decompositions, one of the key attribute is the notion of the rank. 
Even though the rank in TR is a vector, we can assume all ranks to be the same, unlike that for tensor-train case where the intermediate ranks are higher, thus providing a single parameter that can be tuned based on the data and the number of samples available. 
The use of trace operation in the tensor ring structure brings challenges for completion as compared to that for tensor train decomposition. The tensor ring structure is equivalent to a cyclic structure in tensor networks, and understanding this structure can help understand completion for more general tensor networks.   
 In this paper, we propose an alternating minimization algorithm for the  tensor ring completion. For the initialization of the this algorithm, we extend the tensor train approximation algorithm in \cite{oseledets2011tensor} for zero-filled missing data. Further, the different sub-problems in alternating minimization are converted to efficient  least square problems, thus significantly  improving the complexity of each sub-problem. We also analyze the storage and computational complexity of the proposed algorithm.
 
 We  note that, to the best of our knowledge, tensor ring completion has never been investigated for tensor completion, even though tensor ring factorization has been proposed in \cite{Zhao2016}. The different novelties as compared to  \cite{Zhao2016} include the initialization algorithm, exclusion of the normalization of tensor factors, utilizing the structure of the different sub-problems of alternating minimization with incomplete data to convert to least squares based problems, and analysis of storage and computational complexity.


The proposed algorithm is evaluated  on a variety of data sets, including Einstein's image, Extended YaleFace Dataset B, and high speed video.
The results are compared with the tensor train completion algorithms in \cite{grasedyck2015variants,phien2016efficient}, and the additional structure in the tensor ring is shown to  significantly improve the performance as compared to using the TT structure.

The rest of the paper is organized as follows. In section \ref{sec:2} we introduce the basic notation and preliminaries on the TR decomposition. In section \ref{sec:3} we outline the problem statement and propose the main algorithm. We also describe the computational complexity of the proposed algorithm. Following that we test the algorithm extensively against competing methods on a number of real and synthetic data experiments in section \ref{sec:4}. Finally we provide conclusion and future research directions in section \ref{sec:5}. The proofs of Lemmas are provided in the Appendix.

%% file: NotationPreliminaries.tex
\section{Notation \& Preliminaries}
\label{sec:2}

In this paper, vector and matrices are represented by bold face lower case letters $({\bf x,y,z,\cdots})$ and bold face capital letters $({\bf X, Y, Z,\cdots})$ respectively. 
A tensor with order more than two is represented by calligraphic letters $(\bf \mathscr{X}, \mathscr{Y}, \mathscr{Z})$. 
For example, an $n^\text{th}$ order tensor is represented by ${\bf \mathscr{X}} \in \mathbb{R}^{I_1 \times I_2 \times \cdots \times I_n}$, where $I_{i: i=1,2,\cdots, n}$ is the tensor dimension along mode $i$. 
The tensor dimension along mode $i$ could be an expression, where the expression inside $()$ is evaluated as a scalar, e.g. $\mathscr{X}\in \mathbb{R}^{(I_1I_2) \times (I_3I_4)\times (I_5I_6)}$ represents a 3-mode tensor where dimensions along each mode is $I_1I_2$, $I_3I_4$, and $I_5I_6$ respectively. 
An entry inside a tensor $\mathscr{X}$ is represented as $\mathscr{X}(i_1, i_2,\cdots, i_n)$, where $i_{k: k=1,2,.., n}$ is the location index along the $k^{\text{th}}$ mode. 
A colon is applied to represent all the elements of a mode in a tensor,  e.g. $\mathscr{X}(:, i_2,\cdots, i_n)$ represents the fiber along mode $1$ and $\mathscr{X}(:, :, i_3, i_4,\cdots, i_n)$ represents the slice along mode $1$ and mode $2$ and so forth. 
Similar to Hadamard product under matrices case, Hadamard product between tensors is the entry-wise product of the two tensors.  
$\text{vec}(\cdot)$ represents the vectorization of the tensor in the argument. The vectorization is carried out lexicographically over the index set, stacking the elements on top of each other in that order. 
Frobenius norm of a tensor is the same as the vector $\ell_2$ norm of the corresponding tensor after vectorization, e.g. $\|\mathscr{X}\|_F =\|\text{vec}(\mathscr{X})\|_{\ell_2}$. 
$\times$ between matrices is the standard matrix product operation.

\vspace{-.05in}
{\definition (Mode-$i$ unfolding \cite{cichocki2014tensor}) Let $\mathscr{X} \in \mathbb{R}^{I_1 \times \cdots \times I_n}$ be a $n$-mode tensor. Mode-$i$ unfolding of $\mathscr{X}$, denoted as $\mathscr{X}_{[i]}$, matrized the tensor $\mathscr{X}$ by putting the $i^{\text{th}}$ mode in the matrix rows and remaining modes with the original order in the columns such that
\begin{equation}
	\mathscr{X}_{[i]} \in \mathbb{R}^{I_i \times (I_1\cdots I_{i-1}I_{i+1}\cdots I_n)}.
\end{equation}
}
\vspace{-.25in}
{\definition(Left Unfolding and Right Unfolding \cite{holtz2012manifolds}) Let $\mathscr{X} \in \mathbb{R}^{R_{i-1} \times I_i \times R_i}$ be a third order tensor, the left unfolding is the matrix obtained by taking the first two modes indices as rows indices and the third mode indices as column indices such that
\begin{equation}
	{\bf L}(\mathscr{X})  = 	(\mathscr{X}_{[3]})^T\in \mathbb{R}^{(R_{i-1}I_i) \times R_i}. 
\end{equation}
Similarly, the right unfolding gives
\begin{equation}
	{\bf R}(\mathscr{X})  = 	\mathscr{X}_{[1]}\in \mathbb{R}^{R_{i-1} \times (I_iR_i)}. 
\end{equation}
}
\vspace{-.25in}
{\definition(Mode-$i$ canonical matrization \cite{cichocki2014tensor} ) Let  $\mathscr{X}\in\mathbb{R}^{I_1 \times \cdots \times I_n}$ be an $n^{\text{th}}$ order tensor, 
the mode-$i$ canonical matrization gives
\begin{equation}
\mathscr{X}_{<i>} \in \mathbb{R}^{(\prod_{t=1}^i I_t) \times (\prod_{t=i+1}^n I_t)},
\end{equation}
such that any entry in $\mathscr{X}_{<i>}$ satisfies
\begin{equation}
\begin{split}
&\mathscr{X}_{<i>}( i_1 + (i_2-1)I_1 +\cdots+ (i_k-1) \prod_{t=1}^{k-1}I_t, \\
&i_{k+1}+(i_{k+2}-1)I_{k+1} + \cdots + (i_n-1) \prod_{t=k+1}^{n-1}I_t)\\
=&\mathscr{X}(i_1,\cdots, i_n) .
\end{split}
\end{equation}
\enddefinition}
{\definition(Tensor Ring \cite{Zhao2016}) Let $\mathscr{X} \in \mathbb{R}^{I_1 \times \cdots \times I_n}$ be a $n$-order tensor with $I_i$-dimension along the $i_\text{th}$ mode, then any entry inside the tensor, denoted as $\mathscr{X}(i_1, \cdots, i_n)$, is represented by
\begin{equation}\label{eq: TR_Sum}
\begin{split}
\mathscr{X}(i_1, \cdots, i_n) = \sum_{r_1=1}^{R_1}\cdots \sum_{r_n=1}^{R_n} &\mathscr{U}_1(r_n, i_1, r_1) \cdots \\
 &\mathscr{U}_n(r_{n-1}, i_n, r_n),
\end{split}
\end{equation}
where $\mathscr{U}_i \in \mathbb{R}^{R_{i-1} \times I_i \times R_i}$ is a set 3-order tensors, also named matrix product states (MPS), that consist the bases of the tensor ring structures. Note that $\mathscr{U}_j(:, i_j, :) \in \mathbb{R}^{R_{j-1} \times 1 \times  R_j}$ can be regarded as a matrix of size $\mathbb{R}^{R_{j-1} \times R_j}$, thus \eqref{eq: TR_Sum} is equivalent to
\begin{equation}\label{eq: TR_Trace}
\mathscr{X}(i_1, \cdots, i_n) = \text{tr}(\mathscr{U}_1(:, i_1, :) \times \cdots \times  \mathscr{U}_n(:, i_n, :)).
\end{equation}
}
{\remark (Tensor Ring Rank (TR-Rank))
In the formulation of tensor ring, we note that tensor ring rank is the vector $[R_1, \cdots, R_n]$. 
In general, $R_{i}$'s are not necessary to be the same. 
In our set-up, motived by the fact that $R_{i}$ and $R_{i-1}$ represent the connection between $\mathscr{U}_i$ with the remaining $\mathscr{U}_{j:j\neq i}$, we set $R_i = R$ $\forall i=1, \cdots, n$, and {\bf the scalar $R$ is referred to as the tensor ring rank} in the remainder of this paper.}
{\remark (Tensor Train \cite{oseledets2011tensor}) Tensor train is a special case of tensor ring when $R_n = 1$. 
}

Based on the formulation of tensor ring structure, we define a tensor connect product, the operation between the MPSs, to describe the generation of high order tensor $\mathscr{X}$ from the sets of MPSs $\mathscr{U}_{i:i=1,\cdots, n}$. Let $R_0\triangleq R_n$ for ease of expressions. 
{\definition(Tensor Connect Product)
Let $\mathscr{U}_i \in \mathbb{R}^{R_{i-1} \times I_i \times R_i}, i=1,\cdots, n$ be $n$ $3$rd-order tensors, the tensor connect product between $\mathscr{U}_j$ and $\mathscr{U}_{j+1}$ is defined as,
\begin{equation}
\begin{split}
\mathscr{U}_j\mathscr{U}_{j+1} & \in \mathbb{R}^{R_{j-1} \times (I_jI_{j+1}) \times R_{j+1}}\\
&= \text{reshape}\left( {\bf L}(\mathscr{U}_j) \times {\bf R}(\mathscr{U}_{j+1}) \right).
\end{split}
\end{equation}
Thus, the tensor connect product $n$ MPSs is   
\begin{equation}
\mathscr{U} =\mathscr{U}_1 \cdots \mathscr{U}_n \in \mathbb{R}^{R_0 \times (I_1\cdots I_n) \times R_n}.
\end{equation}
}
Tensor connect product gives the product rule for the production between $3$-order tensors, just like the matrix product as for $2$-order tensor. Under matrix case, $\mathscr{U}_j \in \mathbb{R}^{1 \times I_j \times R_j}$, $\mathscr{U}_{j+1} \in \mathbb{R}^{R_j \times I_{j+1} \times 1}$. Thus tensor connect product gives the vectorized solution of matrix product. 

We then define an operator $f$ that applies on $\mathscr{U}$.  Let $\mathscr{U} \in \mathbb{R}^{R_0 \times (I_1 \cdots I_n) \times R_n}$ be the  $3$-order tensor, $R_0 = R_n$, and let $f$ be a reshaping operator function that reshapes a $3$-order tensor $\mathscr{U}$ to a tensor of dimension $\mathscr{X}$ of dimension $\mathbb{R}^{I_1 \times \cdots \times I_n} $, denoted as 
\vspace{-.05in}
\begin{equation}
\mathscr{X} = f(\mathscr{U}),
\end{equation}
\vspace{-.05in}
where $\mathscr{X}(i_1,\cdots, i_n)$  is generated by 
\begin{equation}
\mathscr{X}(i_1,\cdots, i_n) = \text{tr}(\mathscr{U}(:, i_1 + (i_2-1)I_1+\cdots + (i_n-1)I_{n-1}, :) ).
\end{equation}

Thus a tensor $\mathscr{X}\in\mathbb{R}^{I_1 \times \cdots \times I_n}$ with tensor ring structure is equivalent to
\vspace{-.05in}
\begin{equation}
\mathscr{X} = f(\mathscr{U}_1 \cdots \mathscr{U}_n).
\end{equation}
\vspace{-.05in}
Similar to matrix transpose, which can be regarded as an operation that cyclic swaps the two modes for a $2$-order tensor, 
we define a `tensor permutation' to describe the cyclic permutation of the tensor modes for a higher order tensor. 
{\definition (Tensor Permutation) 
For any $n$-order tensor $\mathscr{X} \in \mathbb{R}^{I_1 \times \cdots \times I_n}$, the $i^{\text{th}} $ tensor permutation is defined as $\mathscr{X}^{P_i} \in \mathbb{R}^{I_i \times I_{i+1} \times \cdots \times I_n \times I_1 \times I_2 \times \cdots \times I_{i-1}}$  such that $\forall_i, j_i\in [1,I_i]$
\begin{equation}\label{eq: TensorPermutation}
\mathscr{X}^{P_i}(j_i,\cdots, j_n,j_1,\cdots, j_{i-1}) = \mathscr{X}(j_1,\cdots,j_n ).
\end{equation}
\enddefinition}

Then we have the following result.
\vspace{-.05in}
\begin{lemma} \label{lemma1} If $\mathscr{X} = f(\mathscr{U}_1 \cdots \mathscr{U}_n)$, then $\mathscr{X}^{P_i} =f( \mathscr{U}_i \mathscr{U}_{i+1} \cdots \mathscr{U}_n \mathscr{U}_1 \cdots \mathscr{U}_{i-1})$.
\end{lemma} 
\vspace{-.05in}
With this background and basic constructs, we now outline the main problem setup.   

%% file: Algorithm.tex
\section{Formulation and Algorithm for Tensor Ring Completion}\label{sec:3}
\subsection{Problem Formulation} 
Given a tensor $\mathscr{X} \in \mathbb{R}^{I_1 \times\cdots \times I_n}$ that is partially observed at locations $\Omega$, let $\mathscr{P}_\Omega \in \mathbb{R}^{I_1 \times\cdots \times I_n}$ be the corresponding binary tensor in which $1$ represents an observed entry and $0$ represents a missing entry.
The problem is to find a low tensor ring rank (TR-Rank) approximation of the tensor $\mathscr{X}$, denoted as $f(\mathscr{U}_1 \cdots \mathscr{U}_n)$,  such that the recovered tensor matches $\mathscr{X}$ at $\mathscr{P}_\Omega$. This problem is referred as the tensor completion problem under tensor ring model,  which is equivalent to the following problem
\vspace{-.05in}
\begin{equation}\label{eq: T0}
\min_{\mathscr{U}_{i:i=1,\cdots, n}} \| \mathscr{P}_\Omega \circ ( f(\mathscr{U}_1 \cdots \mathscr{U}_n) -\mathscr{X}) \|_F^2 .
\end{equation}
Note that the rank of the tensor ring $R$ is predefined and the dimension of $\mathscr{U}_{i:i=1,\cdots, n}$ is $\mathbb{R}^{R \times I_i \times R}$.

To solve this problem, we propose an algorithm, referred as Tensor Ring completion by Alternating Least Square (TR-ALS) to solve the problem in two steps. 
\begin{itemize}
\vspace{-.1in}
\item Choose an initial starting point by using Tensor Ring Approximation (TRA).  This initialization algorithm is detailed in Section \ref{initial}.
\vspace{-.1in}
\item Update the solution by applying Alternating Least Square (ALS) that alternatively (in a cyclic order) estimates a factor say $\mathscr{U}_{i}$ keeping the other factors fixed. This algorithm is detailed in Section \ref{leastsquare}.
\end{itemize}

\subsection{Tensor Ring Approximation (TRA)}\label{initial}
A heuristic initialization algorithm, namely TRA, for solving \eqref{eq: T0} is proposed in this section. The proposed algorithm is a modified version of tensor train decomposition as proposed in  \cite{oseledets2011tensor}. We first perform a tensor train decomposition on the zero-filled data, where the rank is constrained by Singular Value Decomposition (SVD). Then,  an approximation for the tensor ring  is formed by extending the obtained factors  to the desired dimensions by filling the remaining entries with small random numbers.  We note that the small entries show faster convergence as compared to zero entries based on our considered small examples, and thus motivates the choice in the algorithm. Further, non-zero random entries help the algorithm initialize with larger ranks since the TT decomposition has the corner ranks as 1. Having non-zero entries can help the algorithm not getting stuck in a local optima of low corner rank. The TRA algorithm is given in Algorithm \ref{Algo_Decomposition}.



\input{Algo_Decomposition}
\subsection{Alternating Least Square}\label{leastsquare}
The proposed tensor ring completion by alternating least square method (TR-ALS) solves \eqref{eq: T0} by solving  the following problem for each $i$ iteratively. The factors are initialized from the TRA algorithm presented in the previous section.
\begin{equation} \label{eq: T2}
\mathscr{U}_i = \argmin_{\mathscr{Y}} \| \mathscr{P}_\Omega \circ f(\mathscr{U}_1 \cdots \mathscr{U}_{i-1}\mathscr{Y}\mathscr{U}_{i+1}\cdots \mathscr{U}_n) -\mathscr{X}_\Omega) \|_F^2.
\end{equation} 
{\lemma \label{lemma2}
When $i \neq 1$, solving 
\begin{equation} \label{eq: tran1}
\mathscr{U}_i = \argmin_{\mathscr{Y}} \| \mathscr{P}_\Omega \circ f(\mathscr{U}_1 \cdots \mathscr{U}_{i-1}\mathscr{Y}\mathscr{U}_{i+1}\cdots \mathscr{U}_n) -\mathscr{X}_\Omega) \|_F^2
\end{equation}
is equivalent to
\begin{equation}\label{eq: transform}
\mathscr{U}_i = \argmin_{\mathscr{Y}} \| \mathscr{P}^{P_i}_\Omega \circ f(\mathscr{Y} \mathscr{U}_{i+1}\cdots \mathscr{U}_n\mathscr{U}_{1}\cdots\mathscr{U}_{i-1}) -\mathscr{X}^{P_i}_\Omega  \|_F^2.
\end{equation} 
}


Since the format of \eqref{eq: transform} is exactly the same for each $i$ when the other factors are known, it is enough to describe solving a single $\mathscr{U}_k$ without loss of generality. Based on Lemma \ref{lemma2}, we need to solve the following problem.
\begin{equation}\label{eq: problemk}
\mathscr{U}_k = \argmin_{\mathscr{Y}} \| \mathscr{P}^{P_k}_\Omega \circ f(\mathscr{Y} \mathscr{U}_{k+1}\cdots \mathscr{U}_n\mathscr{U}_{1}\cdots\mathscr{U}_{k-1}) -\mathscr{X}^{P_k}_\Omega  \|_F^2.
\end{equation} 

We further apply mode-$k$ unfolding, which gives the equivalent problem
\begin{equation}\label{eq: problem1}
\begin{split}
\mathscr{U}_k 
&= \argmin_{\mathscr{Y}} \| {\mathscr{P}^{P_k}_\Omega}_{[k]} \circ {f(\mathscr{Y} \mathscr{U}_{k+1}\cdots \mathscr{U}_n\mathscr{U}_{1}\cdots\mathscr{U}_{k-1})}_{[k]} \\
&-{\mathscr{X}^{P_k}_\Omega}_{[k]}  \|_F^2,
\end{split}
\end{equation} 
where ${\mathscr{P}^{P_k}_{\Omega}}_{[k]}$, $ {f(\mathscr{Y} \mathscr{U}_{k+1}\cdots \mathscr{U}_n\mathscr{U}_{1}\cdots\mathscr{U}_{k-1})}_{[k]}$ and ${\mathscr{X}^{P_k}_\Omega}_{[k]} $ are matrices with dimension $\mathbb{R}^{I_k \times (I_{k+1}\cdots I_nI_1 \cdots I_{k-1})}$.

The trick in solving \eqref{eq: problem1} is that each slice of tensor $\mathscr{Y}$, denoted as $\mathscr{Y}(:, i_k, :), i_k\in \{1,\cdots, I_k\}$ which corresponds to each row of ${\mathscr{P}^{P_k}_{\Omega}}_{[k]}$, $ {f(\mathscr{Y} \mathscr{U}_{k+1}\cdots \mathscr{U}_n\mathscr{U}_{1}\cdots\mathscr{U}_{k-1})}_{[k]}$ and ${\mathscr{X}^{P_k}_\Omega}_{[k]} $,   can be solved independently, thus equation \eqref{eq: problem1} can be solved by solving $I_k$ equivalent subproblems 
\begin{equation}\label{eq: T4}
\begin{split}
&\mathscr{U}_k(:, i_k, :)= \argmin_{\mathscr {Z} \in \mathbb{R}^{R \times 1 \times R}}\\
& \|{\mathscr{P}^{P_k}_\Omega}_{[k]}(i_k, :) \circ f( \mathscr{Z}\mathscr{U}_{k+1}\cdots \mathscr{U}_{k-1})
- {\mathscr{X}^{P_k}_\Omega}_{[k]}(i_k, :)\|_F^2.
\end{split}
\end{equation}

Let 
$\mathscr{B}^{(k)} = \mathscr{U}_{k+1} \cdots \mathscr{U}_n \mathscr{U}_1 \cdots \mathscr{U}_{k-1} \in \mathbb{R}^{R \times (I_{k+1} \cdots I_nI_1 \cdots I_{k-1}) \times R}$, 
$\Omega_{i_k}$ be the observed entries in vector $\mathscr{X}_{[k]}(i_k, :)$, 
thus 
$\mathscr{B}^{(k)}_{\Omega_{i_k}} \in \mathbb{R}^{R \times (I_{k+1} \cdots I_nI_1 \cdots I_{k-1})_{\Omega_{i_k}} \times R}$ are the components in $\mathscr{B}^{(k)}$ 
such that ${\mathscr{P}^{P_k}_\Omega}_{[k]}(i_k, (I_{k+1} \cdots I_nI_1 \cdots I_{k-1})_{\Omega_{i_k}})$ are observed.
Thus equation \eqref{eq: T4}   is equivalent to
\begin{equation}\label{eq: T5}
\begin{split}
\mathscr{U}_k(:, i_k,:) =&\argmin_{\mathscr{Z}} \| f(\mathscr{Z}\mathscr{B}^{(k)}_{\Omega_{i_k}}) \\
&- {\mathscr{X}^{P_k}_\Omega}_{[k]}(i_k, (I_{k+1} \cdots I_nI_1 \cdots I_{k-1})_{\Omega_{i_k}})) \|_F^2.
\end{split}
\end{equation}

We regard $\mathscr{Z} \in \mathbb{R}^{R \times 1 \times R}$ as a matrix ${\bf Z} \in \mathbb{R}^{R \times R}$. Since the Frobenius norm of a vector in \eqref{eq: T5} is equivalent to entry-wise square summation of all entries, we rewrite \eqref{eq: T5} as 
\begin{equation}\label{eq: T5-1}
\begin{split}
&\mathscr{U}_k(:, i_k,:) =\argmin_{{\bf Z}\in \mathbb{R}^{R \times R}} \\
&\sum_{j \in \Omega_{i_k}} \| \text{tr}({\bf Z} \times \mathscr{B}^{(k)}_{\Omega_{i_k}}(:, j, :))
 -  {\mathscr{X}^{P_k}_\Omega}_{[k]}(i_k, j)\|_F^2.
\end{split}
\end{equation}

{\lemma \label{lemma3} Let ${\bf A} \in \mathbb{R}^{r_1 \times r_2}$ and ${\bf B} \in \mathbb{R}^{r_2 \times r_1}$ be any two matrices, then 
\begin{equation}
\begin{split}
\text{Trace}({\bf A} \times {\bf B}) &=   vec({\bf B}^\top)^\top vec({\bf A}).
\end{split}
\end{equation}
}

Based on Lemma \ref{lemma3}, \eqref{eq: T5-1} becomes
\begin{equation}\label{eq: T5-2}
\begin{split}
&\mathscr{U}_k(:, i_k,:) =\argmin_{\bf Z} \sum_{j \in \Omega^{(k)}_{i_k}} \\
 &\| \text{vec}((\mathscr{B}^{(k)}_{\Omega_{i_k}}(:, j, :))^\top)^\top \text{vec}({\bf Z})
-  {\mathscr{X}^{P_k}_\Omega}_{[k]}(i_k, j)\|_F^2.
\end{split}
\end{equation}

Then the problem for solving $\mathscr{U}_k[:, i_k, :]$ becomes a least square problem. Solving $I_k$ least square problem would give the optimal solution for $\mathscr{U}_k$.
Since each $\mathscr{U}_{i:i=1,\cdots,n}$ can solved by a least square method, tensor completion under tensor ring model can be solved by taking orders to update $\mathscr{U}_{i: i=1,\cdots, n}$ until convergence. 
We note the completion algorithm does not require normalization on each MPS, unlike the decomposition algorithm \cite{Zhao2016} that normalizes all the MPSs to seek a unique factorization. The stopping criteria in TR-ALS is measured via the changes of the last tensor factors $\mathscr{U}_n$ since if  the last factor does not change, the other factors are less likely to change. 
Details of the algorithm are given in Algorithm \ref{main_algo}. 
\input{Algo_LeastSquare}

%% file: Algo_Decomposition.tex
\begin{algorithm}[t!]
   \caption{Tensor Ring Approximation (TRA)}
   \label{Algo_Decomposition}
   \begin{algorithmic}[1]
   \INPUT  Missing entry zero filled tensor $\mathscr{X} \in \mathbb{R}^{I_1\times I_2\times \cdots \times I_n}$, TR-Rank $R$, small random variable depicting the standard deviation of the added normal random variable $\sigma$
   \OUTPUT Tensor train decomposition $\mathscr{U}_{i: i=1,\cdots, n} \in \mathbb{R}^{R\times I_i \times R}$
   
   \STATE Apply mode-1 canonical matricization for $\mathscr{X}$ and get matrix ${\bf X}_1 =\mathscr{X}_{<1>} \in \mathbb{R}^{I_1 \times (I_2 I_3 \cdots I_n)}$
    
    \STATE Apply SVD and threshold the number of singular values to be $T_1 = \text{min}(R, I_1, I_2\cdots I_n)$, such that ${\bf X}_1 = {\bf U}_1 {\bf S}_1 {\bf V}_1^\top, 
    {\bf U}_1 \in \mathbb{R}^{I_1 \times T_1}, {\bf S}_1 \in \mathbb{R}^{T_1 \times T_1}, {\bf V}_1 \in \mathbb{R}^{T_1 \times (I_2 I_3 \cdots I_n)}$.  
    Reshape ${\bf U}_1$ to $\mathbb{R}^{1\times I_1 \times T_1}$ and extend it to $\mathscr{U}_1 \in \mathbb{R}^{R \times I_1 \times R}$ by filling the extended entries by random normal distributed values sampled from $\mathcal{N}(0,\sigma^2)$. 
    \STATE Let ${\bf M}_1 ={\bf S}_1 {\bf V}_1^\top \in \mathbb{R}^{T_1 \times ({I_2 I_3 \cdots I_n}) }$.

	\FOR{$i=2 $ to $n-1$}
	\STATE Reshape ${\bf M}_{i-1}$ to ${\bf X}_i \in \mathbb{R}^{(T_{i-1} I_i) \times (I_{i+1} I_{i+2} \cdots I_{n})}$.
	\STATE Compute SVD and threshold the number of singular values to be $T_i = min(R, T_{i-1}I_i, I_{i+1}\cdots I_n)$, such that ${\bf X}_{i} = {\bf U}_i {\bf S}_i {\bf V}_i^\top, 
	{\bf U}_i \in \mathbb{R}^{(T_{i-1}  I_i) \times T_i}, {\bf S}_i \in \mathbb{R}^{T_i \times T_i}, {\bf V} \in \mathbb{R}^{T_i \times (I_{i+1} I_{i+2} \cdots I_n)} $. 
	Reshape ${\bf U}_i$ to $\mathbb{R}^{T_{i-1} \times I_i \times T_i}$ and extend it to $\mathscr{U}_i \in \mathbb{R}^{R \times I_i \times R}$ by filling the extended entries by random normal distributed values sampled from $\mathcal{N}(0,\sigma^2)$. 
	\STATE Set ${\bf M}_i ={\bf S}_i {\bf V}_i^\top \in \mathbb{R}^{T_i \times (I_{i+1} I_{i+2} \cdots  I_n)}$
	\ENDFOR
	
	\STATE Reshape ${\bf M}_{n-1} \in \mathbb{R}^{T_{n-1} \times I_n}$ to $\mathbb{R}^{T_{n-1} \times I_n \times 1}$, and  extend it to $\mathscr{U}_n \in \mathbb{R}^{R \times I_n \times R}$ by filling the extended entries by random normal distributed values sampled from $\mathcal{N}(0,\sigma^2)$ to get $\mathscr{U}_n$
	\STATE Return $\mathscr{U}_1,\cdots, \mathscr{U}_n$
\end{algorithmic}
\label{main_algo}
\end{algorithm} 

%% file: Algo_LeastSquare.tex
\begin{algorithm}[t!]
   \caption{TR-ALS Algorithm}
   \begin{algorithmic}[1]
   \INPUT  Zero-filled Tensor $\mathscr{X}_\Omega \in \mathbb{R}^{I_1\times I_2\times ... \times I_n}$, binary observation index tensor $\mathscr{P}_\Omega \in \mathbb{R}^{I_1\times I_2\times ... \times I_n}$, tensor ring rank $R$, thresholding parameter $tot$, maximum iteration $maxiter$ 
    \OUTPUT Recovered tensor $\mathscr{X}_R$
    \STATE       Apply tensor ring approximation in Algorithm 1 on $\mathscr{X}_\Omega$ to initialize the MPSs $\mathscr{U}_{i:i=1,\cdots, n} \in\mathbb{R}^{R\times I_i \times R}$. Set iteration parameter $\ell = 0$. 
	\WHILE{ $\ell \leq maxiter$ }
	\STATE $\ell = \ell +1$
	\FOR {$i=1$ to $n$}
	\STATE \text{\bf Solve by Least Square Method} 
	${\mathscr{U}_i}^{(\ell)} = \argmin_\mathscr{U} \|\mathscr{P}_\Omega \circ (\mathscr{U} \mathscr{U}_{i+1}^{(\ell-1)}...\mathscr{U}_{n}^{(\ell-1)}\mathscr{U}_{1}^{(\ell)}...\mathscr{U}_{i-1}^{(\ell)}    - \mathscr{X})\|_F^2$ 
	\ENDFOR	
	\IF {$\frac{ \|\mathscr{U}_n^{(\ell+1)} -\mathscr{U}_n^{(\ell)} \|_F}{ \| {\mathscr{U}}^{(\ell)}_n\|_F} \leq tot$}
	\STATE Break
	\ENDIF
	\ENDWHILE
	\STATE Return $\mathscr{X}_R = \text{reshape}(\mathscr{U}_1^{(\ell)} \mathscr{U}_2^{(\ell)}...\mathscr{U}_{n-1}^{(\ell)} \mathscr{U}_n^{(\ell)})$
\end{algorithmic}
\label{main_algo}
\end{algorithm}

%% file: Complexity.tex
\subsection{Complexity Analysis}
{\bf Storage Complexity} Given  an $n$-order tensor $\mathscr{X} \in \mathbb{R}^{I_1 \times \cdots  \times I_n}$, the total amount of parameters to store is $\prod_{i=1}^n I_i$, which increases exponentially with order. Under tensor ring model, we can reduce the storage space by converting each factor (except the last) one by one to being orthonormal and multiply the product with the next factor. Thus, the number of parameters to store the MPSs $\mathscr{U}_{i:i=1,\cdots, n-1}$ with orthonormal property requires storage $\sum_{i=1}^{n-1} (R^{2}I_i - R^2)$, and $\mathscr{U}_n$ with parameter $R^2I_n$. Thus, the total amount of storage is $R^2(\sum_i^nI_i -n+1)$, where the tensor ring rank $R$ can be adjusted to fit the tensor data at the desired accuracy. 


{\bf Computational Complexity}  For each $\mathscr{U}_i$, the least square problem in \eqref{eq: problem1} solved by pseudo-inverse gives a computational complexity $\max(O(P R^4), O(R^6) )$, where $P$ is the total number of observations. Within one iteration when $n$ MPSs need to be updated, the overall complexity is $\max(O(nPR^4), O(nR^6) )$.

We note that tensor train completion \cite{grasedyck2015variants} gives the similar complexity as tensor ring completion. 
However, tensor train rank is a vector and it is hard for tuning to achieve the optimal completion.  The intermediate ranks in tensor train are large in general, leading to significantly higher computational complexity of tensor train. This is alleviated in part by the tensor ring structure which can be parametrized by the tensor ring rank which can be smaller than the intermediate ranks of the tensor train in general. In addition, the single parameter in the tensor ring structure leads to an ease in characterizing the performance for different ranks and can be easily tuned for practical applications. The lower ranks lead to lower computational complexity of data completion under the tensor ring structure as compared to the tensor train structure. 


%% file: Experiment.tex
\section{Numerical Results}\label{sec:4}
In this section, we  compare our proposed TR-ALS algorithm with tensor train completion under alternating least square (TT-ALS) algorithm \cite{grasedyck2015variants}, which solves the tensor completion by alternating least squares under tensor train format. 
SiLRTC algorithm is another tensor train completion algorithm proposed in \cite{phien2016efficient} and the tensor train rank is tuned based on the dimensionality of the tensor. It is selected for comparison as it shows good recovery in image completion \cite{phien2016efficient}.
The evaluation merit we consider is Recovery Error (RE). Let $\hat{\mathscr{X}}$ be the recovered tensor and $\mathscr{X}$ be the ground truth of the tensor. Thus, the recovery error is defined as
\begin{equation*}
RE = \frac{\|\hat{\mathscr{X}} - \mathscr{X}\|_F}{\|\mathscr{X}\|_F}.
\end{equation*}
Tensor ring completion by alternating least square (TR-ALS ) algorithm is an iterative algorithm and  the maximum iteration, $maxiter$, is set to be 300. 
The convergence is captured by the change of the last factorization term $\mathscr{U}_n$, where the error tolerance is set to be $10^{-10}$.

In the remaining of the section, we first evaluate the completion results for synthetic data. Then we validate the proposed TR-ALS algorithm on image completion, YaleFace image-sets completion, and video completion.

\subsection{Synthetic Data}
In this section, we consider a completion problem of a $4$-order tensor $\mathscr{X}\in\mathbb{R}^{20 \times 20 \times 20 \times 20}$ with TR-Rank being $8$ without loss of generality. The tensor is generated by a sequence of connected $3$-rd order tensor $\mathscr{U}_{i: i=1,\cdots, 4} \in \mathbb{R}^{8 \times 20 \times 8}$ and every entry in $\mathscr{U}_i$ are sampled independently from a standard normal distribution.

TT-ALS is considered as a comparable to show the difference between tensor train model and tensor ring model. Two different tensor train ranks are chosen for the comparisons. The first tensor-train  ranks are chosen as  $[8, 8, 8]$, and the completion with these ranks is called Low rank tensor train (LR-TT) completion. The second tensor-train  ranks are chosen as the double of the first (  $[16, 16, 16]$), and the completion with these ranks is called High rank tensor train (HR-TT) completion.  Another comparable used is the SiLRTC algorithm proposed in \cite{phien2016efficient}, where the  rank is adjusted according to the dimensionality of the tensor data, and a heuristic factor of  $f=1$ in the proposed algorithm of \cite{phien2016efficient} is selected for testing.

\begin{figure*}[t!]
	\centering
	\begin{subfigure}[b]{0.5\textwidth}
		\centering
		\includegraphics[trim=.45in 1.5in .8in 2.8in, clip, width = 0.8\textwidth]{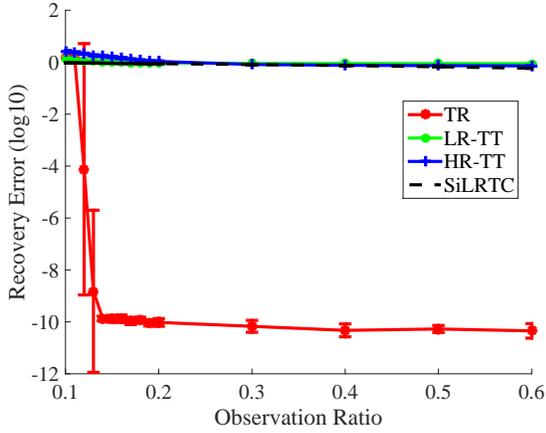}\vspace{-.4in}
		\caption{\small Recovery error versus observation Ratio. Average of $10$ experiments for TR-ALS with TR-Rank $8$, TT-ALS with TT-Rank $[8,8,8]$, TT-ALS with TT-Rank $[16,16,16]$, and SiLRTC are shown for comparison. Error bar marked using one standard deviation.}
		\label{Synthetic}
	\end{subfigure}%
	~ 
	\begin{subfigure}[b]{0.5\textwidth}
		\centering
		\includegraphics[trim=.45in 1.5in .8in 2.8in, clip, width = 0.8\textwidth]{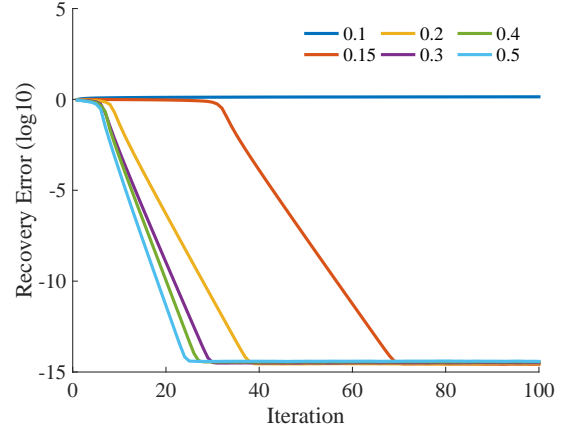}\vspace{-.2in}
		\caption{Convergence plot for TR-ALS under observation ratio being from 0.1 to 0.5 for $4_\text{th}$ order tensor of dimension $20\times 20\times 20\times 20$ with TR-Rank $8$.  }
		\label{Convergence}
	\end{subfigure}%
	~
	\vspace{-.1in}
	\caption{\small Synthetic data is a $4_\text{th}$ order tensor of dimension $20\times 20\times 20\times 20$ with TR-Rank being $8$.}
	\vspace{-.1in}
\end{figure*}

Fig.\ref{Synthetic} shows the completion error of TR-ALS, LR-TT, HR-TT, and SiLRTC for observation ratio from $10\%$ to $60\%$. TR-ALS shows the lowest recovery error compared with other algorithms and the recovery error drops to $10^{-10}$ for observation ratio larger than $14\%$.
The large completion errors of all tensor train algorithm at every observation ratio show that tensor train algorithm can not effectively complete the tensor data generated under tensor ring model. 
Fig. \ref{Convergence} shows the convergence of TR-ALS under sampling ratios $10\%, 15\%, 20\%, 30\%, 40\%, \text{ and } 50\%$, and the plot indicates the higher the observation ratios, the faster the algorithm converges. 
When the observation ratio is lower than $10\%$, the tensor with missing data can not be completed under the proposed set-up. 
The fast convergence of the proposed TR-ALS algorithm indicates that alternating least square is effective in tensor ring completion. 
\vspace{-.1in}
\subsection{Image Completion}
In this section, we consider the completion of RGB Einstein Image \cite{Einstein_Web}, treated as a $3$-order tensor $\mathscr{X} \in \mathbb{R}^{600 \times 600 \times 3}$. A reshaping operation is applied to transform the image into a $7$-order tensor of size $\mathbb{R}^{6\times 10 \times 10 \times 6 \times 10 \times 10 \times 3}$. 
Reshaping low order tensors into high order tensors is a common practice in literature and has shown improved performance in classification \cite{novikov2015tensorizing} and completion \cite{phien2016efficient}.
\begin{figure*}[t!]
	\centering
	\begin{subfigure}[b]{0.5\textwidth}
		\centering
		\includegraphics[trim=.45in 1.5in .8in 2.8in, clip, width = 0.9\textwidth]{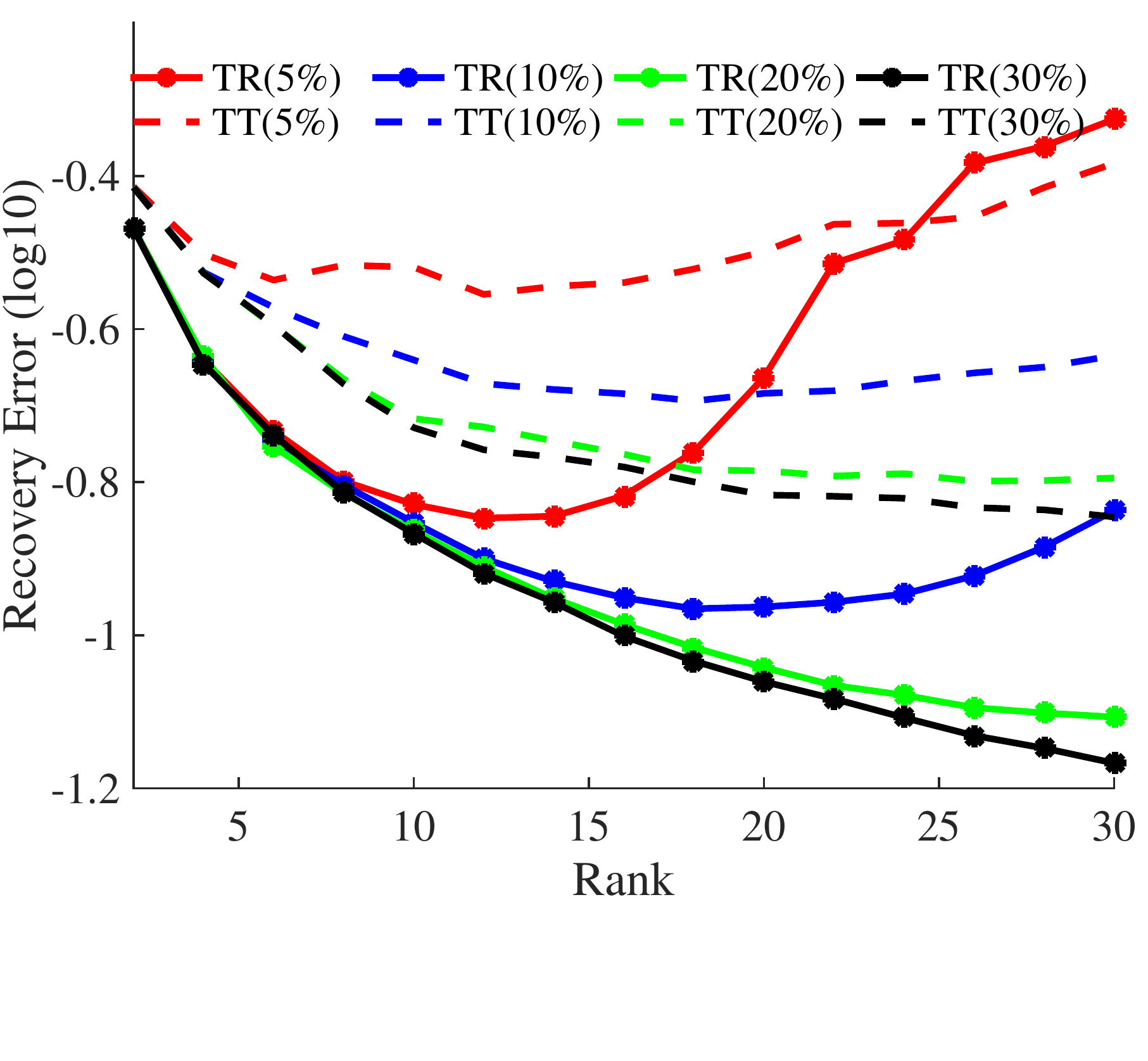}\vspace{-.4in}
		\caption{The recovery error versus rank for TR-ALS and TT-ALS under observation ratio $5\%, 10\%, 20\%, 30\%$}
		\label{Einstein}
	\end{subfigure}%
	~ 
	\begin{subfigure}[b]{0.5\textwidth}
		\centering
		\includegraphics[trim= 0in 0in 0in 0in, clip, width = \textwidth]{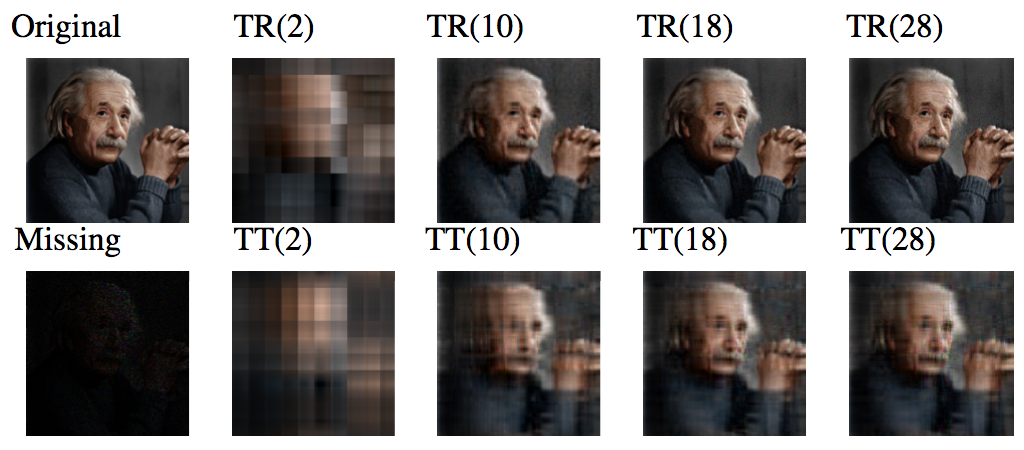}\vspace{.0in}
		\caption{The completed Einstein Image when 10\% of pixels are randomly observed. The first column is the original Einstein image and Einstein image with 10\% randomly observed entries. 
In the remaining 4 columns, the first row are completed images by TR-ALS with TR-Rank $2,10,18, 28$ and the second row are images completed by TT-ALS with the same TT-Rank. The completion errors for TR-ALS and TT-ALS are $33.97\%, 14.03\%, {\bf 10.83\%}, 14.55\%$ and $38.51\%, 22.89\%, {\bf 20.70\%},  23.19\%$ accordingly.}
		\label{Einstein10}
	\end{subfigure}%
	\vspace{-0.1in}
	\caption{\small Einstein image is of size $600 \times 600 \times 3$, and is further reshaped into a $7$-order tensor of size $6\times 10 \times 10 \times 6 \times 10 \times 10 \times 3$ tensor for tensor ring completion}\vspace{-.0in}
\end{figure*}

Fig. \ref{Einstein} shows the recovery error versus rank for TR-ALS and TT-ALS when the percentage of data observed are  $5\%, 10\%, 20\%, 30\%$.  At any considered ranks, TR-ALS completes the image with a better accuracy than TT-ALS. 
For any given percentage of observations,  the recovery error first decreases as the rank increases which is caused by the increased information being captured by the increased number of parameters in the tensor structure.  
The recovery error then starts to increase after a thresholding rank, which can be ascribed to over-fitting. 
\emph{The plot also indicates that higher the observation ratio, larger the thresholding rank, which to the best of our knowledge is reported for the first time}. Fig. \ref{Einstein10} shows the recovered image of Einstein image when $10\%$ pixels are randomly observed. TR-ALS with rank $28$ gives the best recovery accuracy in the considered ranks.

\vspace{-.1in}
\subsection{YaleFace Dataset Completion}
In this section, we consider Extended YaleFace Dataset B \cite{georghiades1998illumination} that includes 38 people with 9 poses under 64 illumination conditions. Each image has the size of $192 \times 168$, where we down-sample the size of each image to $48 \times 42$ for ease of computation. 
We consider the image subsets of 38 people under 64 illumination with 1 pose by formatting the data into a $4$-order tensor in $\mathbb{R}^{48 \times 42 \times 64 \times 38}$, which is further reshaped into a $8$-order tensor $\mathscr{X}\in \mathbb{R}^{6 \times 8 \times 6 \times 7 \times 8 \times 8 \times 19 \times 2}$. 
We consider the case when $10\%$ of pixels are randomly observed.
YaleFace sets completion is considered to be harder than an image completion since features under different illumination and across human features are harder to learn than information from the color channels of images. 
\begin{table*}[t!]
  \centering
  \scalebox{1}{
  \begin{tabular}{ccccccc}
    \toprule
    Rank		& 5 & 10 & 15 & 20 & 25 & 30\\
    \midrule
    TT-ALS		($\mathbb{R}^{6 \times 8 \times 6 \times 7 \times 8 \times 8 \times 19 \times 2}$)
    			&$37.08\% $		&$29.65\%$ 	&$27.91\%$	&$26.84\%$ 	&$26.16\%$ 	&${\ 25.55\%}$ \\
    TR-ALS		($\mathbb{R}^{6 \times 8 \times 6 \times 7 \times 8 \times 8 \times 19 \times 2}$)	
    			&$33.45\% $		&${\bf 24.67\%}$ 	&${\bf 20.72\%}$	&${\bf 18.47\%}$ 	&${\bf 16.92\%}$ 	&${\bf 16.25\%}$ \\
    TR-ALS	($\mathbb{R}^{2\times 3 \times 2 \times 4 \times 2 \times 3 \times 7 \times 8 \times 8 \times 19 \times 2}$)	
    			&$33.73\% $		&$25.08\%$ 	&$21.20\%$	&$18.97\%$ 	&$17.34\%$ 	&${ 16.34\%}$ \\
    TR-ALS	($\mathbb{R}^{48 \times 42 \times 64 \times 38}$)	
    			&${\bf 30.36\%} $		&$26.08\%$ 	&$23.74\%$	&$22.22\%$ 	&$21.48\%$ 	&${21.57  \%}$ \\		
    \bottomrule
  \end{tabular}
  }
  \caption{\small Completion error of 10\% observed Extended YaleFace data for TT-ALS and TR-ALS under rank $5, 10, 15, 20, 25, 30$.   }
  \vspace{-.1in}
  \label{YaleFace_Error}
\end{table*}
Table \ref{YaleFace_Error} shows that for any considered rank, TR-ALS recovers data better than TT-ALS and the best completion result in the given set-up is 16.25\% for TR-ALS as compared with $25.55\%$ given by TT-ALS. 
Further we reshape the data into an $11$-order tensor and $4$-order tensor to evaluate the effect of reshaped tensor size on tensor completion. The result in Table \ref{YaleFace_Error} shows that in the given reshaping set-up, reshaping tensor from $4$-order tensor to $7$-th order tensor significantly improve the performance of tensor completion by decreasing recovery error from $21.48\%$ to $16.25\%$. However, further reshaping to $11$-order tensor slightly degrades the  performance of completion, resulting in an increased recovery error to $16.34\%$.
\begin{figure*}[h]
Original
\includegraphics [trim=.05in .1in 2in .2in, keepaspectratio, width=0.8\textwidth] {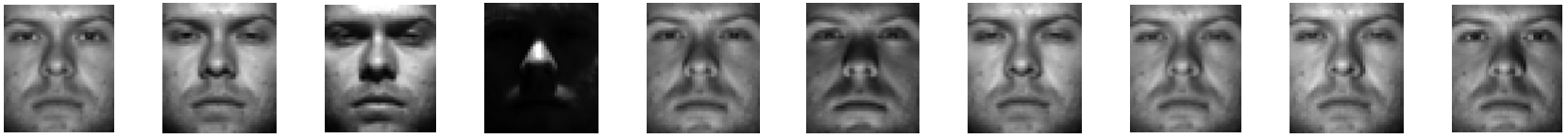}\\
Missing
\includegraphics [trim=.05in .1in 2in .2in, keepaspectratio, width=0.8\textwidth] {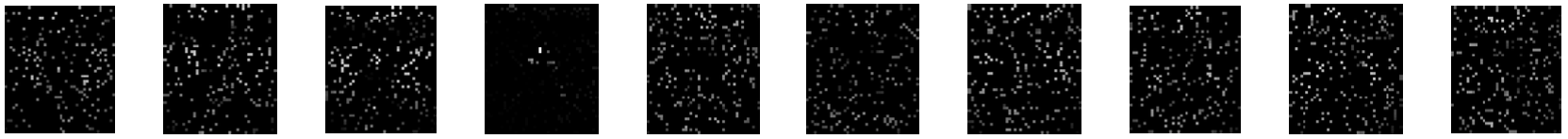}\\
TR(10)
\includegraphics [trim=.05in .1in 2in .2in, keepaspectratio, width=0.8\textwidth] {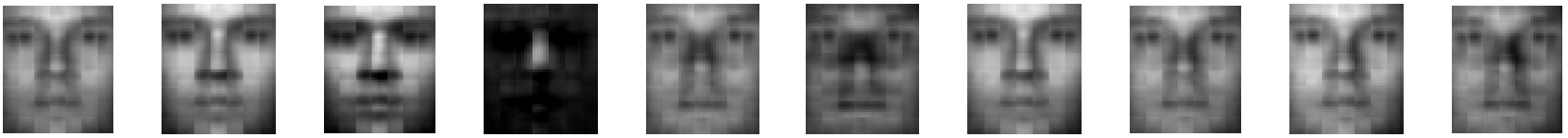}\\
TR(20)
\includegraphics [trim=.05in .1in 2in .2in, keepaspectratio, width=0.8\textwidth] {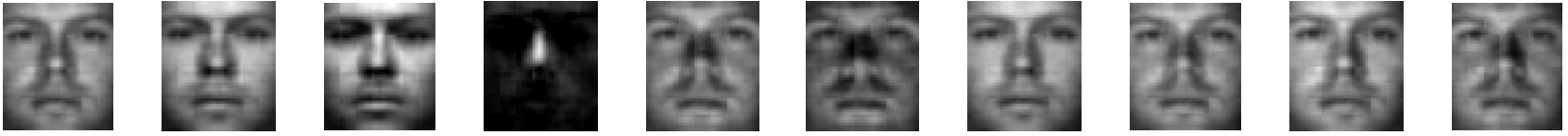}\\
TR(30)
\includegraphics [trim=.05in .1in 2in .2in, keepaspectratio, width=0.8\textwidth] {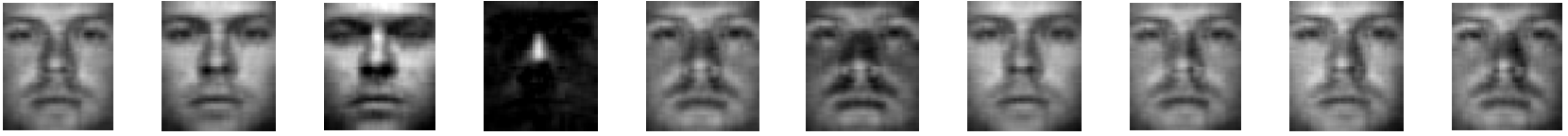}\\
TT(10)
\includegraphics [trim=.05in .1in 2in .2in, keepaspectratio, width=0.8\textwidth] {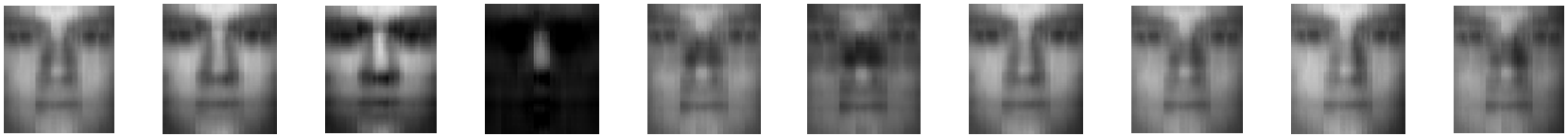}\\
TT(20)
\includegraphics [trim=.05in .1in 2in .2in, keepaspectratio, width=0.8\textwidth] {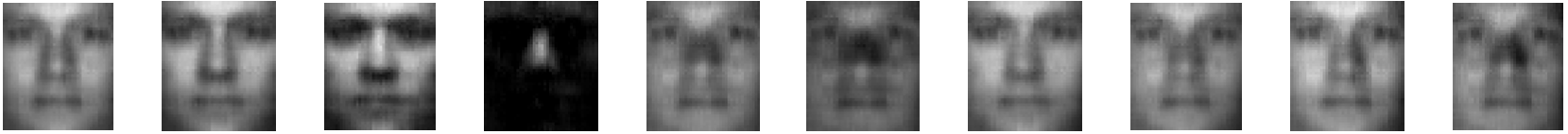}\\
TT(30)
\includegraphics [trim=.05in .1in 2in .2in, keepaspectratio, width=0.8\textwidth] {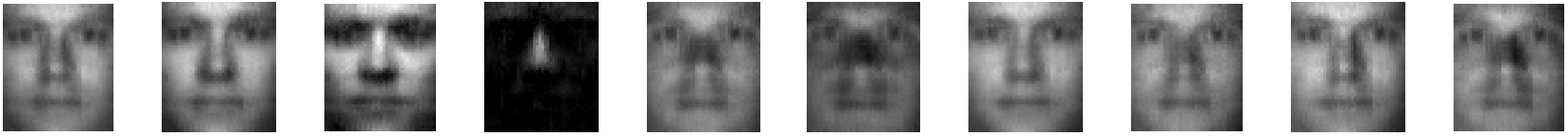}\\
\centering
\vspace{-.2in}
\caption{\small 
YaleFace dataset is sub-sampled to formulated into a tensor of size $\mathbb{R}^{48\times 42 \times 64 \times 38}$, and is further reshaped into a $8$-order tensor of size $  6 \times 8 \times 6 \times 7 \times 8 \times 8 \times 19 \times 2$  for tensor ring completion. $90\%$ of the pixels are assumed to be randomly missing. 
YaleFace dataset completion. From top to bottom are original images, missing images, TR-ALS completed images for TR-Ranks $10, 20, 30$, and TT-ALS completed images for ranks $10, 20, 30$.
}
\label{FaceSets}
\end{figure*}

 Fig. \ref{FaceSets} shows the original image, missing images, and recovered images using TR-ALS and TT-ALS algorithms for ranks of  $10,20, \text{ and } 30$, where the completion results given by TR-ALS better captures the detail information given from the image and recovers the image with a better resolution.

\vspace{-.05in}
\subsection{Video completion}
\vspace{-.05in}
The video data we used in this section is high speed camera video for gun shooting \cite{Video_Web}. It is downloaded from Youtube with 85 frames in total and each frame is consisted by a $100 \times 260 \times 3$ image. 
Thus the video is  a $4$-order tensor of size ${100 \times 260 \times 3 \times 85}$, which is further reshaped into a $11$-order tensor of size ${5 \times 2 \times 5 \times 2 \times 13 \times 2 \times 5 \times 2 \times 3 \times 5 \times 17}$ for completion. 
Video is a multi-dimensional data with different color channel a time dimension  in addition to the 2D image structure. 

\begin{table*}[t!]
  \centering
  \scalebox{1}{
  \begin{tabular}{cccccc}
    \toprule
    Rank		& 10 & 15 & 20 & 25 &30\\
    \midrule
    TT-ALS		&$19.16\% $		&$14.83\%$ 	&$16.42\%$	&$16.86\%$ 	& $16.99\%$ \\
    TR-ALS		&$13.90\% $		&$10.12\%$ 	&$8.13\%$	&$6.88\%$ 		& ${\bf 6.25}\%$\\
    \bottomrule
  \end{tabular}
  }
  \caption{\small Completion error of 10\% observed Video data for TT-ALS and TR-ALS under rank $10, 15, 20, 25$.   }
    \vspace{-.1in}
  \label{Video_Error}
\end{table*}

\begin{figure*}[t!]
Original \hspace{.6in}  TR(10) \hspace{.6in}  TR(15) \hspace{.6in}  TR(20) \hspace{.6in}  TR(25) \hspace{.6in}  TR(30)\\
\includegraphics [trim=.2in .1in -.2in -.1in, keepaspectratio, width=0.15\textwidth] {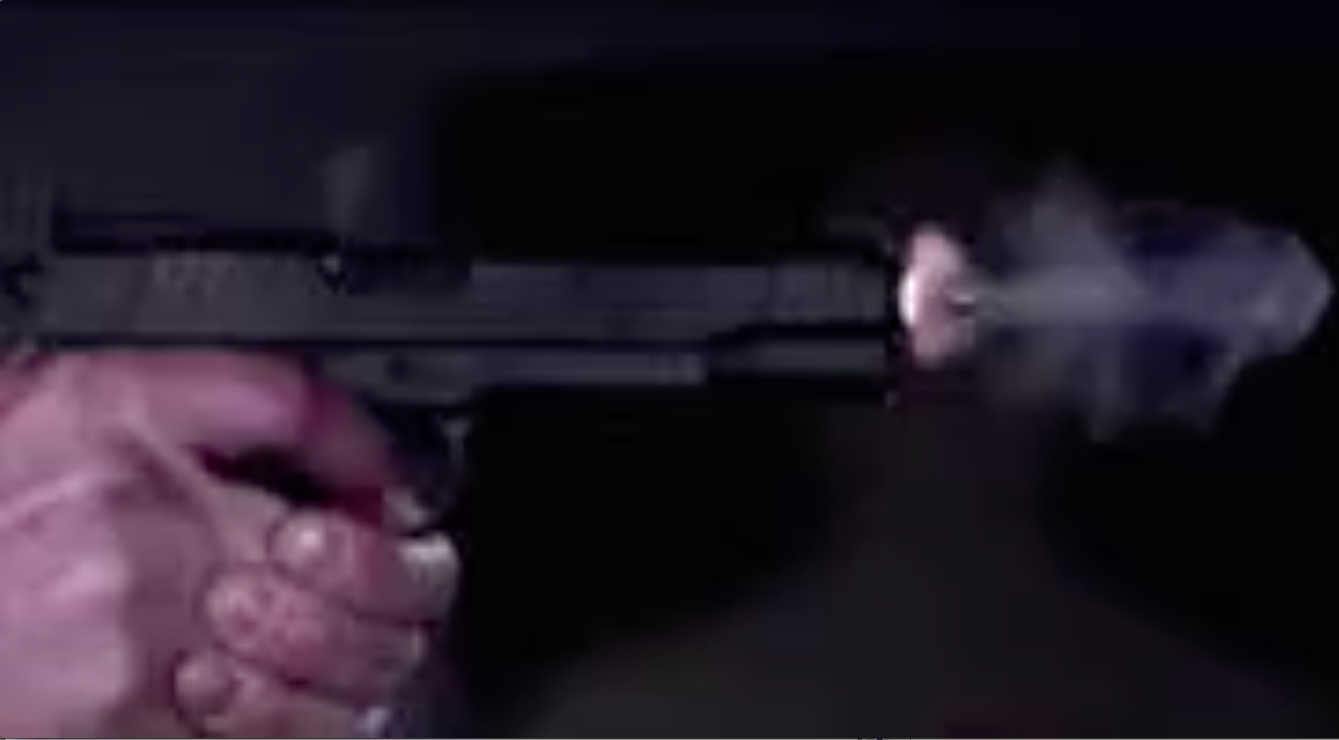}
\includegraphics [trim=.2in .1in -.2in -.1in, keepaspectratio, width=0.15\textwidth] {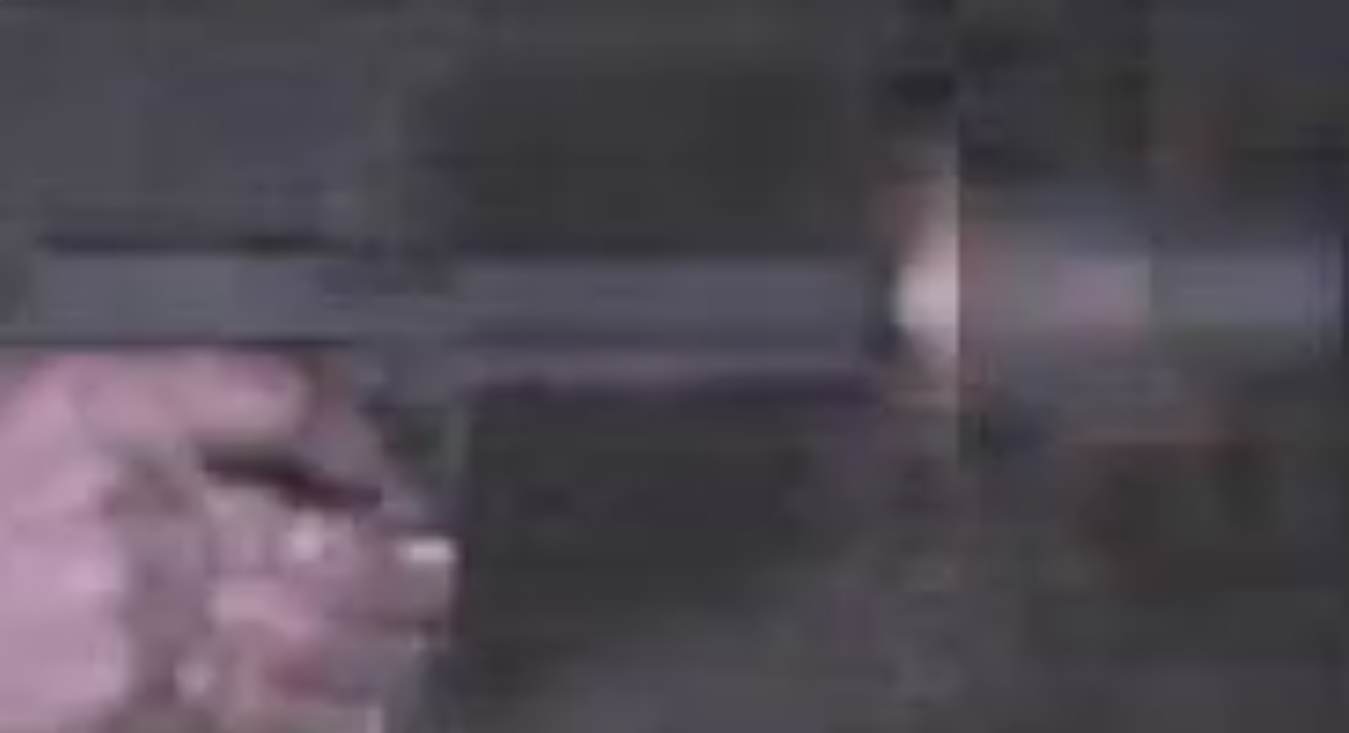}
\includegraphics [trim=.2in .1in -.2in -.1in, keepaspectratio, width=0.15\textwidth] {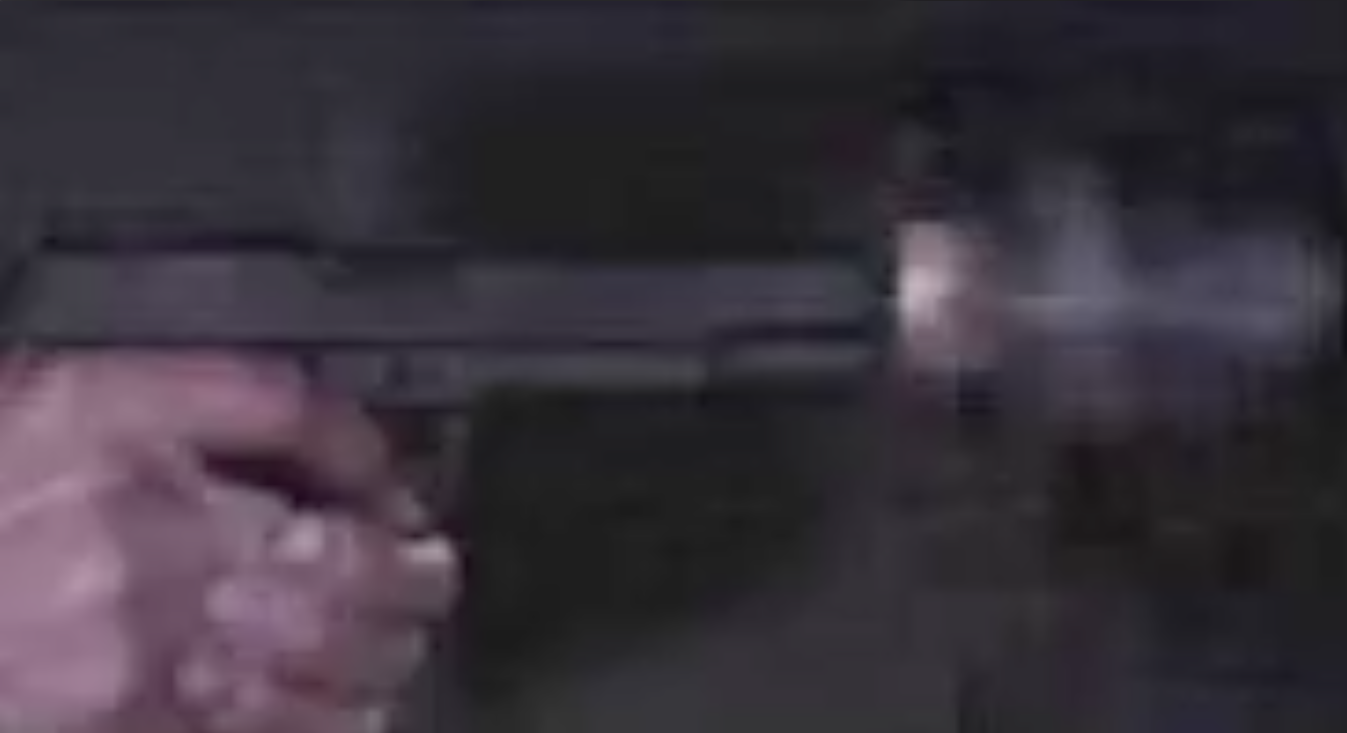}
\includegraphics [trim=.2in .1in -.2in -.1in, keepaspectratio, width=0.15\textwidth] {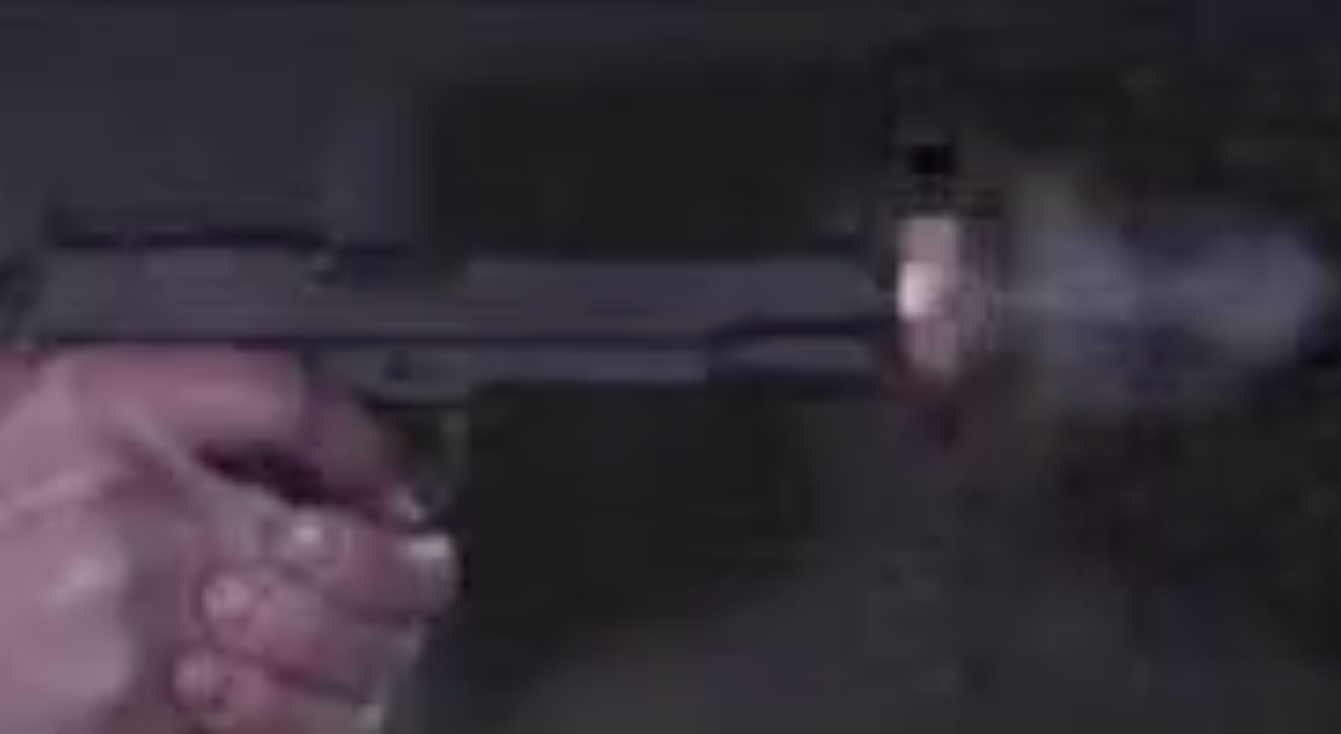}
\includegraphics [trim=.2in .1in -.2in -.1in, keepaspectratio, width=0.15\textwidth] {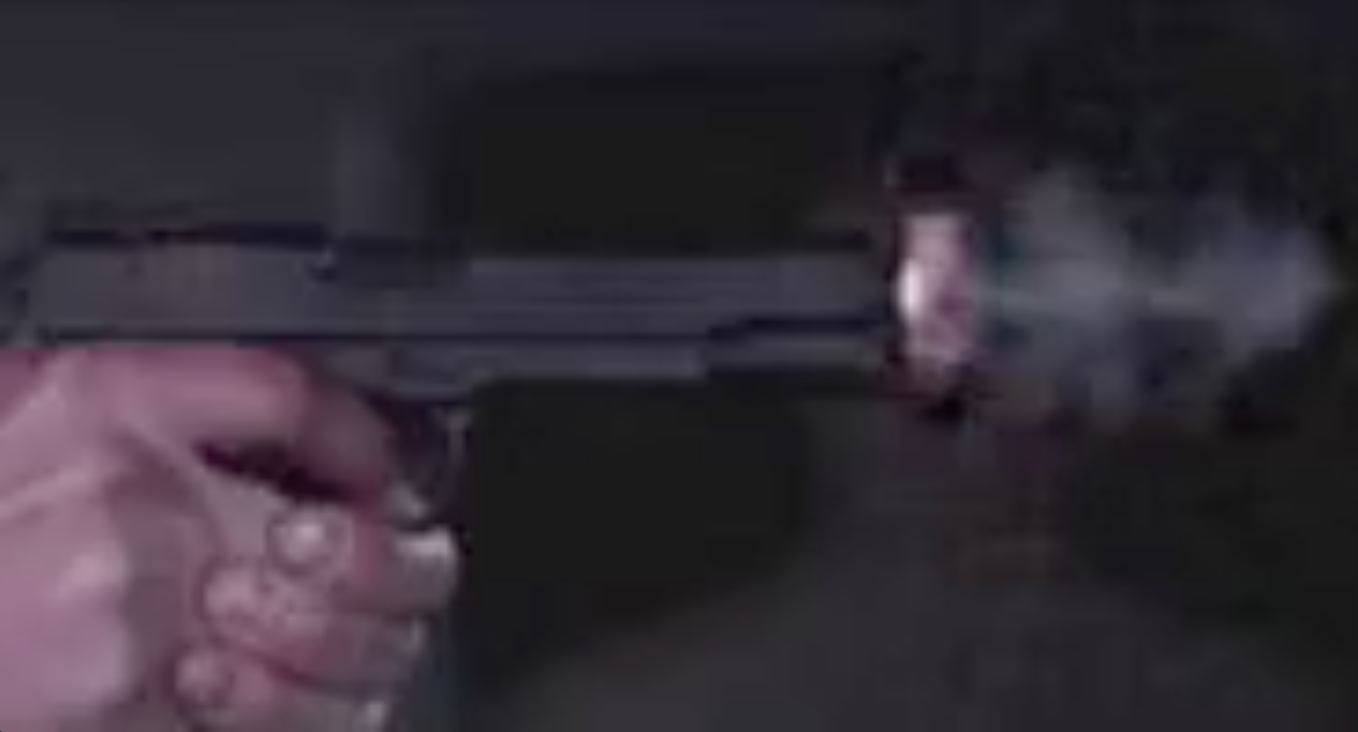}
\includegraphics [trim=.2in .1in -.2in -.1in, keepaspectratio, width=0.15\textwidth] {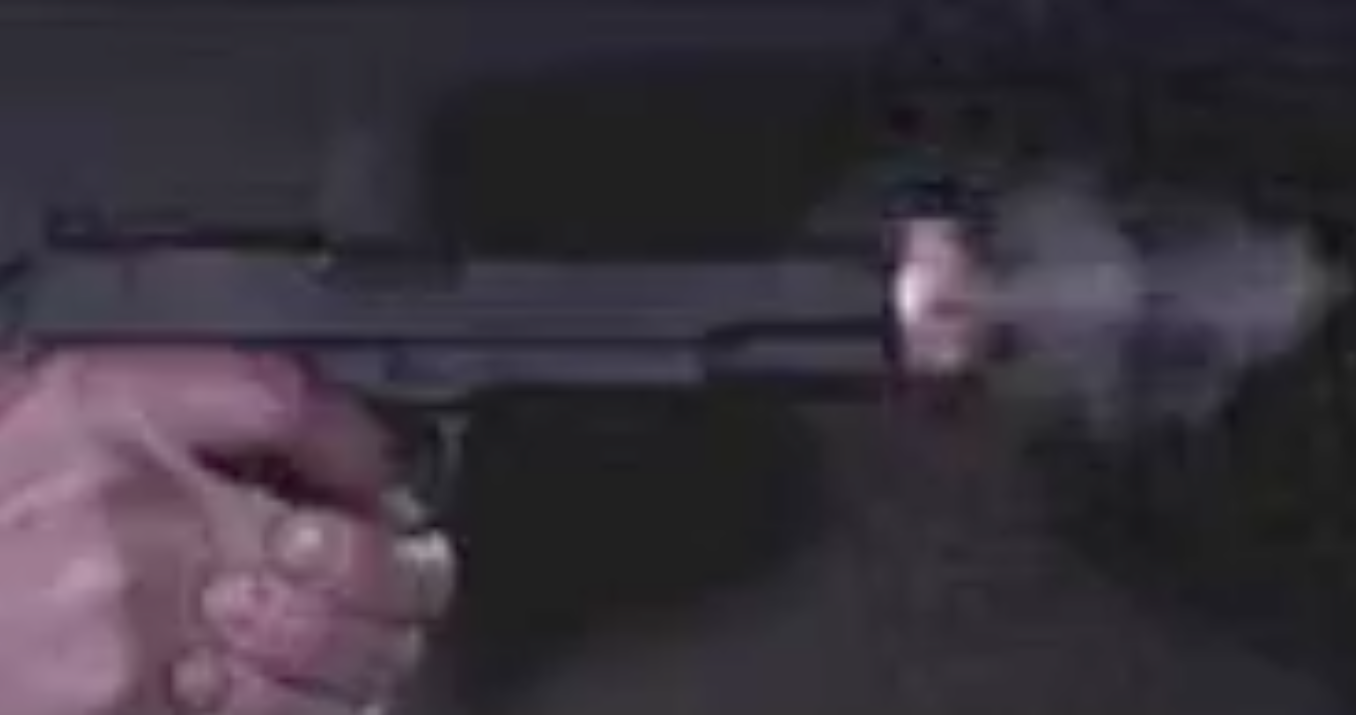}\\
Missing \hspace{.6in}  TT(10) \hspace{.6in}  TT(15) \hspace{.6in}  TT(20) \hspace{.6in}  TT(25) \hspace{.6in}  TT(30)\\
\includegraphics [trim=.2in .1in -.2in -.1in, keepaspectratio, width=0.15\textwidth] {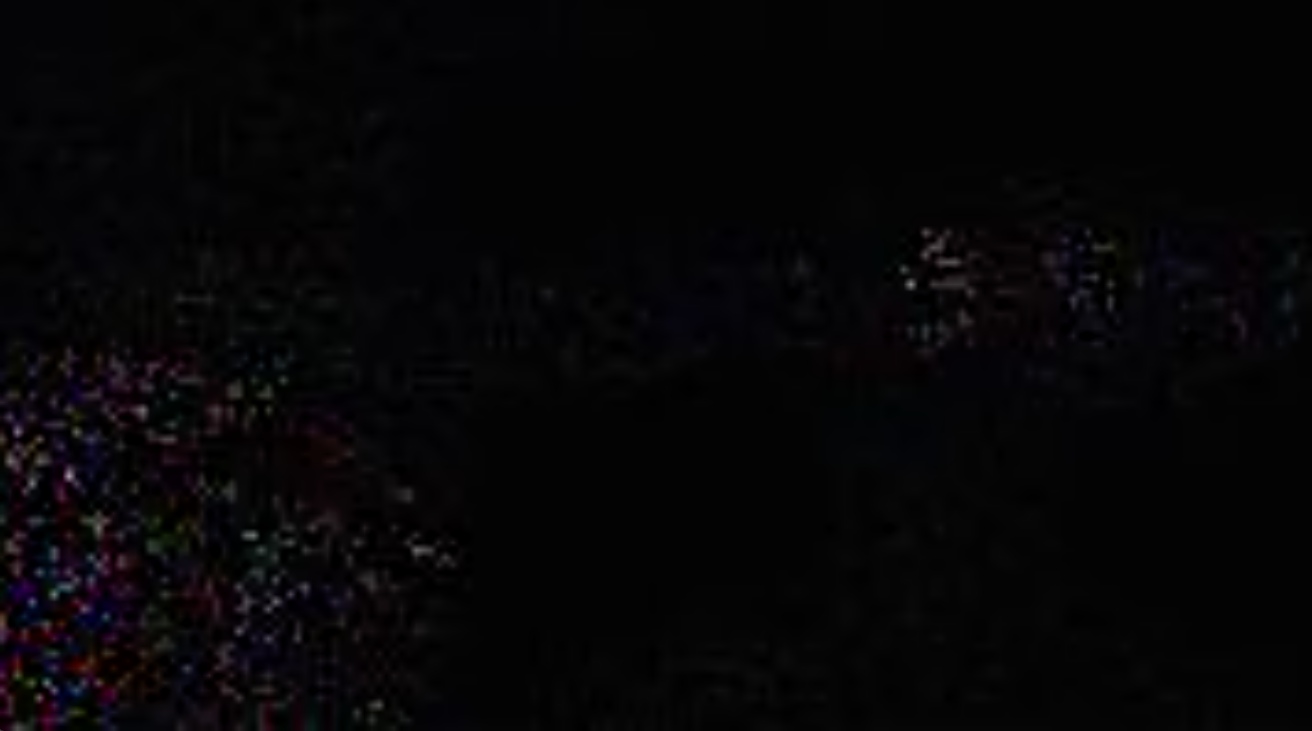}
\includegraphics [trim=.2in .1in -.2in -.1in, keepaspectratio, width=0.15\textwidth] {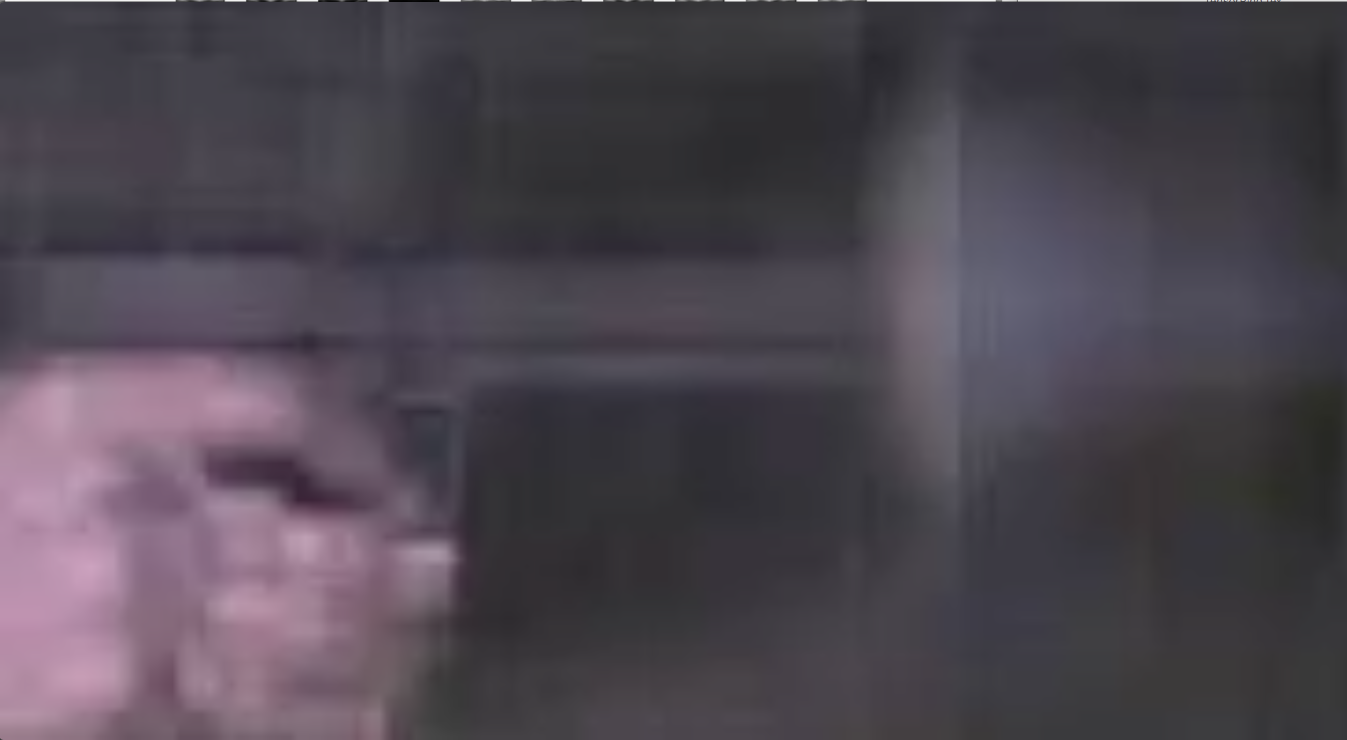}
\includegraphics [trim=.2in .1in -.2in -.1in, keepaspectratio, width=0.15\textwidth] {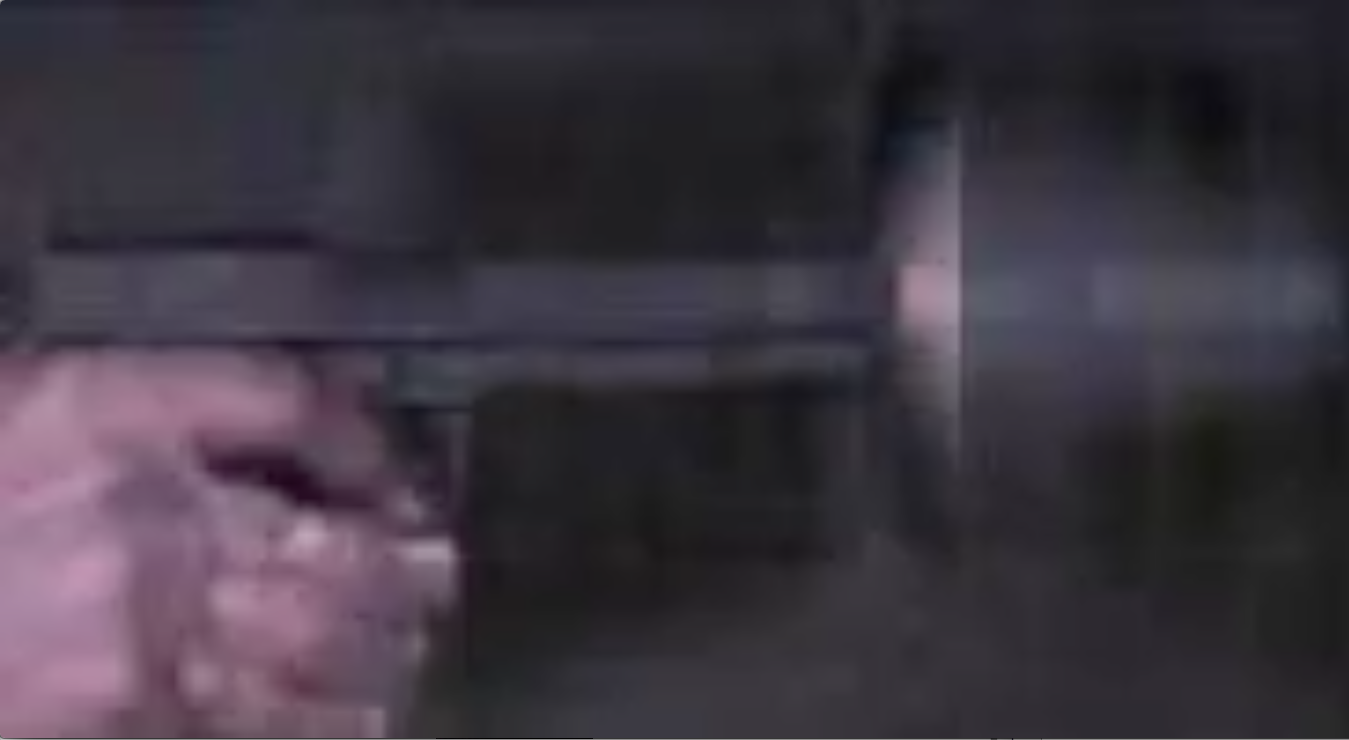}
\includegraphics [trim=.2in .1in -.2in -.1in, keepaspectratio, width=0.15\textwidth] {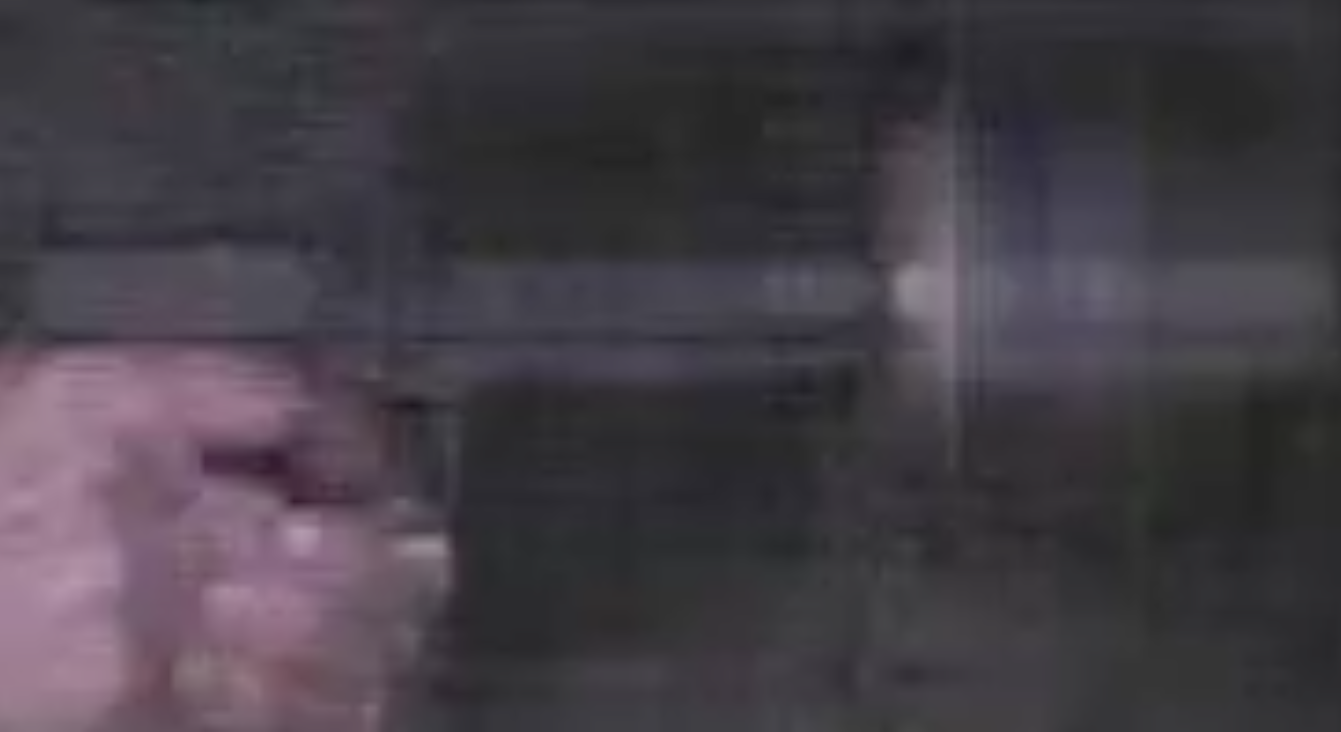}
\includegraphics [trim=.2in .1in -.2in -.1in, keepaspectratio, width=0.15\textwidth] {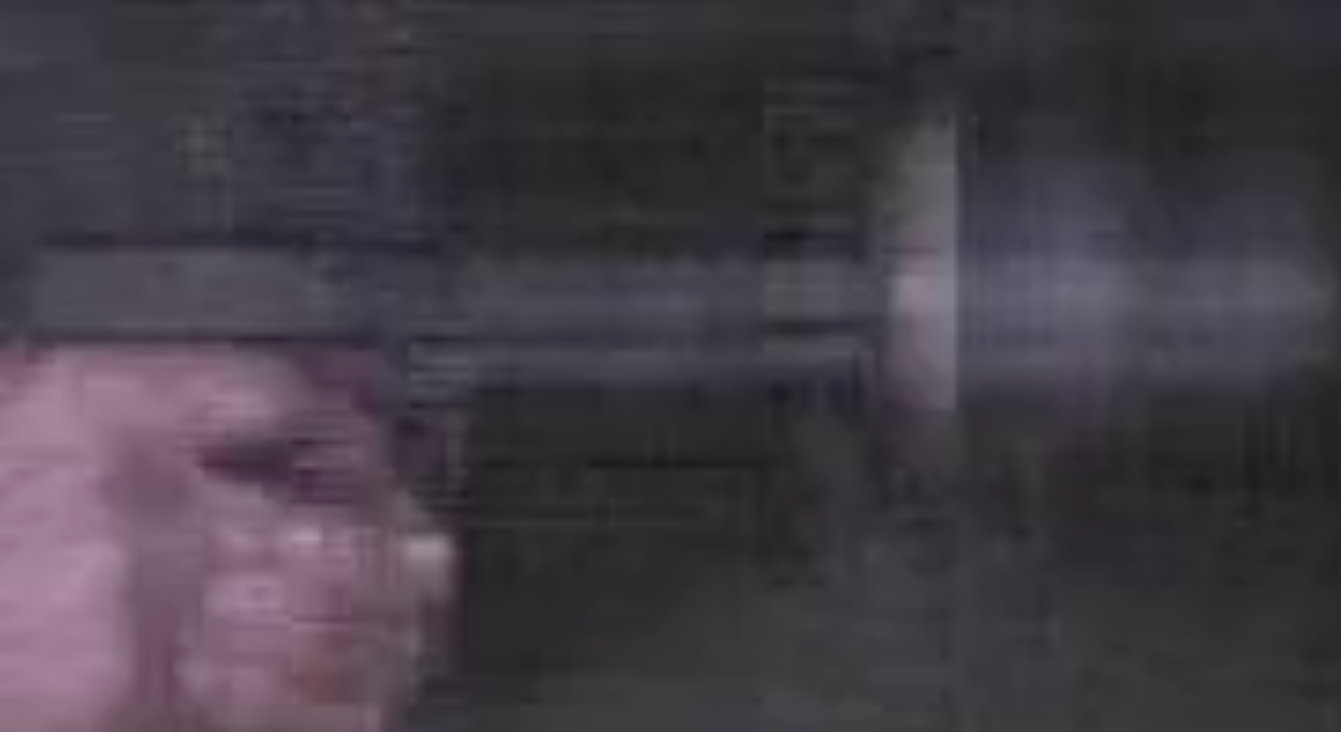}
\includegraphics [trim=.2in .1in -.2in -.1in, keepaspectratio, width=0.15\textwidth] {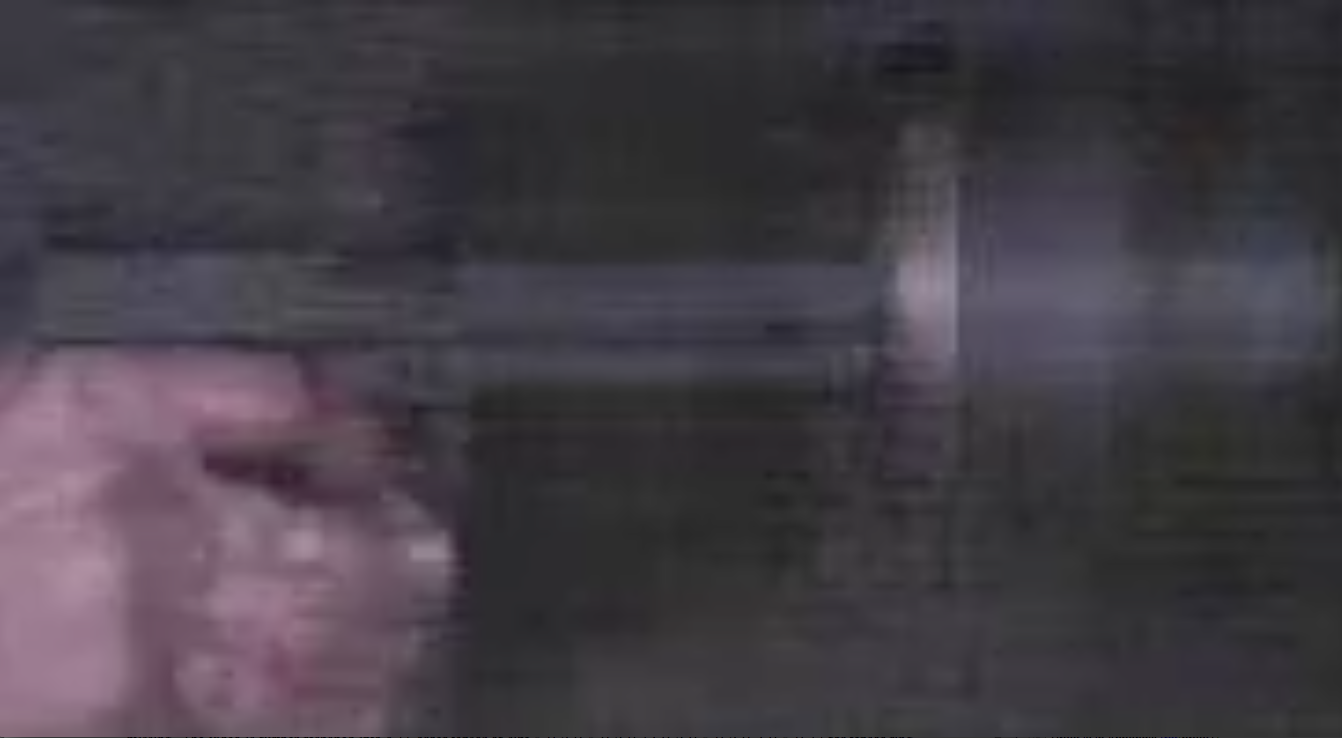}\\
\centering
\vspace{-.2in}
\caption{\small Gun Shot is a video download from Youtube of size $100 \times 260 \times 3 \times 80$.  $90\%$ of the pixels are assumed to be randomly missing. The video is further reshaped into a $11$-order tensor of size $  5\times2\times5\times2\times13\times2\times5\times2\times3\times5\times17$ for tensor ring completion. 
From left to right in the top row, images are the first frame of original video and TR-ALS completed frame under TR-Rank $10,15, 20, 25, 30$. 
From left to right in the second row, images are the first frame of missing video and TT-ALS completed frame under rank $10,15, 20, 25, 30$.
}
\label{Video}
  \vspace{-.15in}
\end{figure*}
In Table \ref{Video_Error}, we show that TR-ALS achieves 6.25\% recovery error when 10\% of the pixels are observed, which is much better than the best recovery error of 14.83\% achieved by TT-ALS.
The first frame of the video is shown in Fig. \ref{Video}, where  the first row shows the original frame and the completed frames by TR-ALS, and  the second row shows the frame with missing entries and the frames completed by TT-ALS.
The resolution, and the display of the bullets and the smoke depict that the proposed TR-ALS achieves better completion results as compared to the TT-ALS algorithm.

%% file: conclusion.tex
\section{Conclusion}\label{sec:5}
We proposed a novel algorithm for data completion using
tensor ring decomposition. This is the first paper on data completion exploiting this structure which is a non-trivial extension of the tensor-train structure.  Our algorithm exploits the matrix product
state representation and uses alternating minimization over the
low rank factors for completion. The proposed approach has been evaluated on a variety of data sets, including Einstein's image, Extended YaleFace Dataset B, and video completion. The evaluation results show significant improvement as compared to the completion using a tensor train decomposition.

Deriving provable performance guarantees on tensor completion
using the proposed algorithm is left as further work. In this context, the statistical
machinery for proving analogous results for the matrix case
\cite{jain2013low, Hardt13a} can be used.





%% file: Apdix.tex
\section{appendix}
\subsection{Proof of Lemma 1}\label{proof0}
\proof
Let ${\bf M} ={\bf M}_1 {\bf M}_2$, thus 
\begin{equation}
{\bf M}(j_1, j_2) =\sum_{j=1}^{r_1} {\bf M}_1(j_1,j) {\bf M}_2(j, j_2)
\end{equation}
where ${\bf M}(j_1, j_2)$ locates at $\text{vec}({\bf M}_1{\bf M}_2)(j_1 + (j_2-1)I_1, 1)$.

Let ${\bf T}_1 \in \mathbb{R}^{(I_1I_2) \times (r_1I_2)} = {\bf I}^{(I_{2})} \otimes {\bf L}(\mathscr{M}_1)$ and ${\bf T}_2\in \mathbb{R}^{(r_1I_{2})  \times 1} = {\bf L}(\mathscr{M}_2)$, and ${\bf T} \in\mathbb{R}^{I_1I_2 \times 1} = {\bf T}_1{\bf T}_2$, thus 
\begin{equation}
\begin{split}
&{\bf T}(j_1+(j_2-1)I_1, 1) \\
= & \sum_{j=1}^{r_1I_2}{\bf T}_1(j_1+(j_2-1)I_1, j) {\bf T}_2(j, 1)\\
= & \sum_{j=(j_2-1)r_1+1}^{j_2r_1} {\bf T}_1(j_1+(j_2-1)I_1, j) {\bf T}_2(j, 1)\\
= & \sum_{j=1}^{r_1}{\bf M}(j_1, j){\bf M}_2(j, j_2)
\end{split}
\end{equation}
We conclude that any ${j_1 + (j_2-1)I_1}^\text{th}$ entry on the left hand side is the same as that on the right hand side, thus we prove our claim. 
\endproof
\subsection{Proof of Lemma 2}\label{proof1}
\proof
Based on definition of tensor permutation in \eqref{eq: TensorPermutation}, on the left hand side, the $(j_1,...., j_n)$ entry of the tensor is 
\begin{equation} \label{eq: prf1}
\mathscr{X}^{P_i}(j_1,...,j_n) = \mathscr{X}(j_{n-i+2},..., j_{n}, j_{1},..., j_{n-i+1}).
\end{equation}

On the right hand side, the $(j_1,...., j_n)$ entry of the tensor gives
\begin{equation}
\begin{split}
& f(\mathscr{U}_i\cdots \mathscr{U}_{i-1})(j_1,\cdots, j_n)\\
=&\text{Trace}(\mathscr{U}_i(:, j_1,:)\mathscr{U}_{i+1}(:, j_2,:)... \mathscr{U}_n(:, j_{n-i+1},:)\\
& \mathscr{U}_1(:, j_{n-i+2}, :) \cdots \mathscr{U}_{i-1}(:, j_n, 1)).
\end{split}
\end{equation}
Since trace is invariant under cyclic permutations, we have
\begin{equation}
\begin{split}
&\text{Trace}(\mathscr{U}_i(:, j_1,:)\mathscr{U}_{i+1}(:, j_2,:)... \mathscr{U}_n(:, j_{n-i+1},:)\\
& \mathscr{U}_1(:, j_{n-i+2}, :) \cdots \mathscr{U}_{i-1}(:, j_n, 1))\\
=&\text{Trace}(
 \mathscr{U}_1(:, j_{n-i+2}, :) \cdots \mathscr{U}_{i-1}(:, j_n, 1)\\
&\mathscr{U}_i(:, j_1,:)\mathscr{U}_{i+1}(:, j_2,:)... \mathscr{U}_n(:, j_{n-i+1},:))\\
= &f(\mathscr{U}_1\cdots \mathscr{U}_{n})(j_{n-i+2},\cdots, j_n, j_1,\cdots, j_{n-i+1}),
\end{split}
\end{equation}
which equals to the right hand side of equation \eqref{eq: prf1}. 
Since any entries in $\mathscr{X}^{P_i}$ are the same as those in $\mathscr{U}_i \mathscr{U}_{i+1} \cdots \mathscr{U}_n \mathscr{U}_1 \cdots \mathscr{U}_{i-1}$, the claim is proved.
\endproof

\subsection{Proof of Lemma 3} \label{proof2}
\proof
First we note that tensor permutation does not change tensor Frobenius norm as all the entries remain the same as those before the permutation. 
Thus, when $i\neq 1$, we permute the tensor inside the Frobenius norm in \eqref{eq: tran1} and get the equivalent equation  as
\begin{equation}\label{eq: p0}
\mathscr{U}_i = \argmin_{\mathscr{Y}} \| \mathscr{P}^{P_i}_\Omega \circ (f(\mathscr{U}_1 \cdots\mathscr{U}_{i-1}\mathscr{Y} \mathscr{U}_{i+1}\cdots \mathscr{U}_n))^{P_i} -\mathscr{X}^{P_i}_\Omega \|_F^2.
\end{equation}

Based on Lemma \ref{lemma1}, we have
\begin{equation}
(f(\mathscr{U}_1 \cdots\mathscr{U}_{i-1}\mathscr{Y} \mathscr{U}_{i+1}\cdots \mathscr{U}_n))^{P_i} =f(\mathscr{Y} \mathscr{U}_{i+1}\cdots \mathscr{U}_n \mathscr{U}_1 \cdots\mathscr{U}_{i-1}),
\end{equation}
thus equation \eqref{eq: p0} becomes 
\begin{equation}\label{eq: p0_1}
\mathscr{U}_i = \argmin_{\mathscr{Y}} \| \mathscr{P}^{P_i}_\Omega \circ f(\mathscr{Y} \mathscr{U}_{i+1}\cdots \mathscr{U}_n \mathscr{U}_1 \cdots\mathscr{U}_{i-1}) -\mathscr{X}^{P_i}_\Omega \|_F^2.
\end{equation}
Comparing \eqref{eq: p0_1} and \eqref{eq: T3}, we have $\mathscr{P}_\Omega,\mathscr{X}_\Omega$ and $\mathscr{U}_2\cdots\mathscr{U}_n $ in \eqref{eq: T3} become $\mathscr{P}_\Omega^{\top_i}, \mathscr{X}_\Omega^{\top_i}$ and $\mathscr{U}_{i+1} \cdots \mathscr{U}_n\mathscr{U}_1 \cdots \mathscr{U}_{i-1}$ in\eqref{eq: p0_1} respectively. Thus we prove our claim. 
\endproof

\subsection{Proof of Lemma 4} \label{proof3}
\proof
\begin{equation}
\begin{split}
\text{Trace}(A \times B) &= \sum_i^{r_1}\left(\sum_j^{r_2} {\bf A}(i,j) {\bf B}(j, i ) \right) \\
&= \sum_i^{r_1}\sum_j^{r_2}  {\bf A}(i,j) {\bf B}^\top(i, j ) \\
& = vec({\bf A})^\top vec({\bf B}^\top)
\end{split}
\end{equation}
\endproof